%% file: ms.tex
\documentclass[final]{cvpr}

\usepackage{times}
\usepackage{epsfig}
\usepackage{graphicx}
\usepackage{amsthm}
\usepackage{amsmath,amssymb} 
\usepackage{color}
\usepackage[inline]{enumitem}
\usepackage{multirow}
\usepackage{colortbl}
\usepackage{booktabs}
\usepackage{balance}
\usepackage{adjustbox}

\usepackage[pagebackref=true,breaklinks=true,colorlinks,bookmarks=false]{hyperref}

\input{macros.tex}

\theoremstyle{definition}
\newtheorem{definition}{Definition}[section]

\begin{document}
	
	\title{On Self-Contact and Human Pose}
	\author{Lea M\"uller$^1$, Ahmed A. A. Osman$^1$, Siyu Tang$^2$, Chun-Hao P. Huang$^1$, Michael J.~Black$^1$\\
		\\
		$^1$Max Planck Institute for Intelligent Systems, T\"ubingen \hspace{0.25in} $^2$ETH Z\"urich\\
		{\tt\small \{lea.mueller,  ahmed.osman, stang, paul.huang, black\}@tuebingen.mpg.de}
	}
		
	\maketitle

	\input{main_sections/00_Abstract.tex}
	
	\input{main_sections/01_Introduction.tex}
	\input{main_sections/02_RelatedWork.tex}

	\input{main_sections/03_SelfContact.tex}
	\input{main_sections/04_TheDataset.tex}

	\input{main_sections/05_Methods.tex}
	\input{main_sections/06_Evaluation.tex}

	\input{main_sections/07_Conclusion.tex}
	
	\textbf{Acknowledgments:} We thank Tsvetelina Alexiadis and  Galina Henz for their help with data collection and Vassilis Choutas for the SMPL-X body measurements and his implementation of Generalized Winding Numbers. The authors thank the International Max Planck Research School for Intelligent Systems (IMPRS-IS) for supporting Lea M\"uller and Ahmed A. A. Osman.
	
	\textbf{Disclosure:}
	MJB has received research gift funds from Adobe, Intel, Nvidia, Facebook, and Amazon. While MJB is a part-time employee of Amazon, his research was performed solely at, and funded solely by, Max Planck. MJB has financial interests in Amazon, Datagen Technologies, and Meshcapade GmbH.

	{\small
		\bibliographystyle{ieee_fullname}
		\balance
		\bibliography{main_supp}
	}

	\newpage
	\appendix
	{\noindent\Large\textbf{Supplementary Material}}
	\newline
	\setcounter{page}{1}
			
	\input{supp_sections/supp_04_TheDataset.tex}
	\input{supp_sections/supp_05_TUCH.tex}

	\input{supp_sections/supp_06_Evaluation.tex}
	\input{supp_sections/supp_06_Evaluation_Qualitative.tex}

\end{document}

%% file: macros.tex
\usepackage{xspace}
\usepackage{xcolor}

\renewcommand{\etal}{et al.\xspace}
\renewcommand{\ie}{i.e.\xspace}
\renewcommand{\eg}{e.g.\xspace}
\newcommand{\suppmat}{Sup.~Mat.\xspace}
\newcommand{\inp}[2]{\langle #1, #2 \rangle}

\newcommand{\geoth}{t_{\mathit{geo}}}
\newcommand{\euclth}{t_{\mathit{eucl}}}

\newcommand{\mtp}{\mbox{MTP}\xspace}

\newcommand{\threedcp}{\mbox{3DCP}\xspace}
\newcommand{\threedcpscan}{\mbox{3DCP Scan}\xspace}
\newcommand{\threedcpmocap}{\mbox{3DCP Mocap.}\xspace}

\newcommand{\dsc}{\mbox{DSC}\xspace}

\newcommand{\nummeshes}{20,114}

\newcommand{\smplifyxmc}{\mbox{SMPLify-XMC}\xspace}
\newcommand{\smplifyxdc}{\mbox{SMPLify-DC}\xspace}

\newcommand{\threedpw}{\mbox{3DPW}\xspace}


%% file: main_sections/00_Abstract.tex
\begin{abstract}
People touch their face 23 times an hour, they cross their arms and legs, put their hands on their hips, etc.  
While many images of people contain some form of self-contact, current 3D human pose and shape (HPS) regression methods typically fail to estimate this contact.
To address this, we develop new datasets and methods that significantly improve human pose estimation with self-contact.
First, we create a dataset of {\em 3D Contact Poses (3DCP)} containing SMPL-X bodies fit to 3D scans as well as poses from AMASS, which we refine to ensure good contact.
Second, we leverage this to create the {\em Mimic-The-Pose (MTP)} dataset of images, collected via Amazon Mechanical Turk, containing people mimicking the 3DCP poses with self-contact.
Third, we develop a novel HPS optimization method, \smplifyxmc, 
that includes contact constraints and uses the known 3DCP body pose during fitting to create near ground-truth poses for MTP images. 
Fourth, for more image variety, we label a dataset of in-the-wild images with {\em Discrete Self-Contact (DSC)} information and use another new optimization method, \smplifyxdc, that exploits discrete contacts during pose optimization.
Finally, we use our datasets during SPIN training to learn a new 3D human pose regressor, called {\em TUCH (Towards
Understanding Contact in Humans)}.
We show that the new self-contact training data significantly improves 3D human pose estimates on withheld test data and existing datasets like 3DPW.  Not only does our method improve results for self-contact poses, but it also improves accuracy for non-contact poses.
The code and data are available for research purposes at \url{https://tuch.is.tue.mpg.de}.

\end{abstract}

%% file: main_sections/01_Introduction.tex
\section{Introduction}

Self-contact takes many forms.  We touch our bodies both consciously and unconsciously~\cite{Kwok:2015}. For the major limbs, contact can provide physical support, whereas we touch our faces in ways that convey our emotional state. We perform self-grooming, we have nervous gestures, and we communicate with each other through combined face and hand motions (e.g.~``shh''). We may wring our hands when worried, cross our arms when defensive, or put our hands behind our head when confident.  
A Google search for ``sitting person'' or ``thinking pose'' for example, will return images, the majority of which, contain self-contact.

\begin{figure}
	\centerline{\includegraphics[width=\columnwidth]{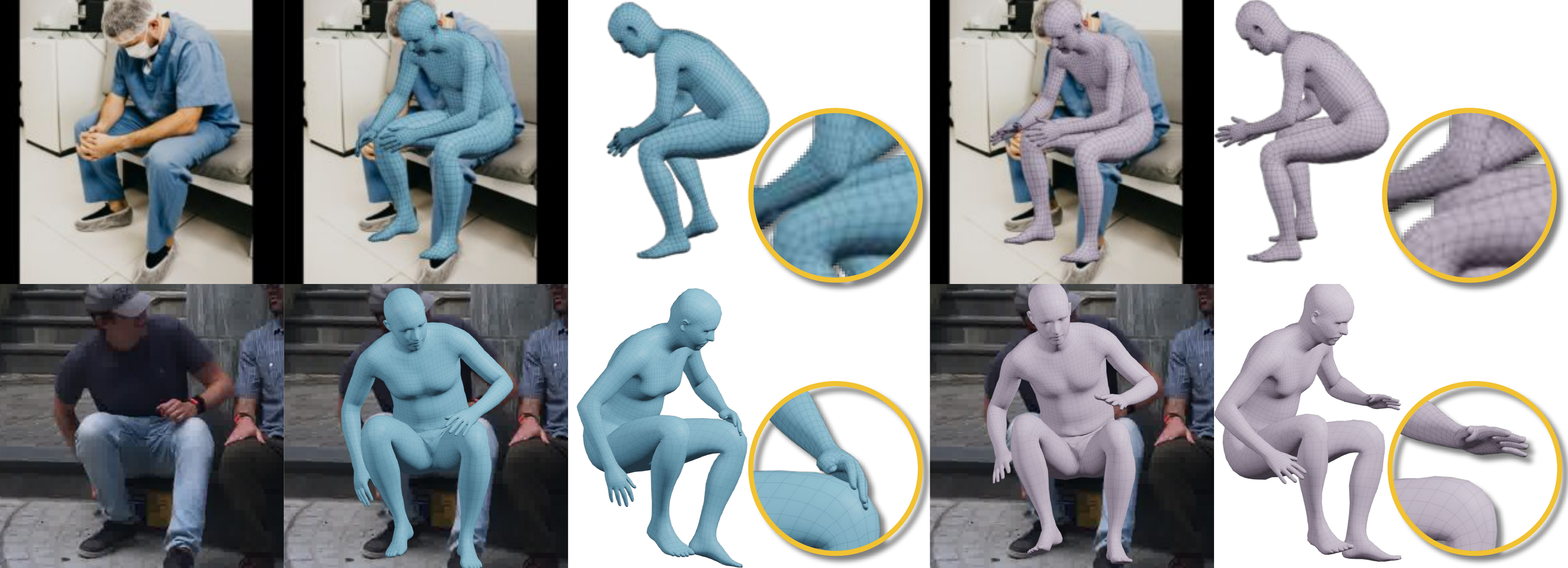}}
	\caption{The first column shows images containing self-contact. In blue (left), results of TUCH, compared to  SPIN results in violet (right). 
	When rendered from the camera view, the estimated pose may look fine (column two vs.~four). 
	However, when rotated, it is clear that training TUCH with self-contact information improves 3D pose estimation (column three vs.~five).}
	\label{fig:rotateview}
\end{figure}

Although self-contact is ubiquitous in human behavior, it is rarely explicitly studied in computer vision. 
For our purposes, self-contact comprises ``self touch'' (where the hands touch the body) and contact between other body parts (e.g.~crossed legs). 
We ignore body parts that are frequently in contact (e.g.~at the crotch or armpits) and focus on contact that is communicative or functional. 
Our goal is to estimate 3D human pose and shape (HPS) accurately for 
any pose.
When self-contact is present, the estimated pose should reflect the true 3D contact.

Unfortunately, existing methods that compute 3D bodies from images perform poorly on images with self-contact; see Fig.~\ref{fig:rotateview}.
Body parts that should be touching generally are not.
Recovering human meshes from images typically involves either learning a regressor from pixels to 3D pose and shape~\cite{kanazawa2018end,kolotouros2019learning}, or fitting a 3D model to image features using an optimization method~\cite{Bogo:ECCV:2016,SMPL-X:2019,xiang2019monocular,xiang2020monoclothcap}. 
The learning approaches rely on labeled training data.
Unfortunately, current 2D datasets typically contain labeled keypoints or segmentation masks but do not provide any information about 3D contact.
Similarly,  existing 3D datasets typically avoid capturing scenarios with self-contact because it complicates mesh processing.
What is missing is a dataset with in-the-wild images and reliable data about 3D self-contact.

To address this limitation, we introduce three new datasets that focus on self-contact at different levels of detail.
Additionally, we introduce two new optimization-based methods that fit 3D bodies to images with contact information.
We leverage these to estimate pseudo ground-truth 3D poses with self-contact.
To make reasoning about contact between body parts, the hands, and the face possible, we represent pose and shape with the SMPL-X~\cite{SMPL-X:2019} body model, which realistically captures the body surface details, including the hands and face.
Our new datasets then let us train neural networks to regress 3D HPS from images of people with self-contact more accurately than state-of-the-art methods.

To begin,  we first construct a {\em 3D Contact Pose (3DCP)} dataset of 3D meshes where body parts are in contact.
We do so using two methods. 
First, we use high-quality 3D scans of subjects performing self-contact poses. We extend previous mesh registration methods to cope with self-contact and register the SMPL-X mesh to the scans. 
To gain more variety of poses, we search the AMASS dataset \cite{AMASS:2019} for poses with self-contact or ``near'' self-contact. We then optimize these poses to bring nearby parts into full contact while resolving interpenetration. This provides a dataset of valid, realistic, self-contact poses in SMPL-X format.

Second, we use these poses to collect a novel dataset of images with near ground-truth 3D pose. 
To do so, we show rendered 3DCP meshes to workers on Amazon Mechanical Turk (AMT).
Their task is to {\em Mimic The Pose (MTP)} as accurately as possible, including the contacts, and submit a photograph.
We then use the ``true'' pose as a strong prior and optimize the pose in the image by extending SMPLify-X \cite{SMPL-X:2019} to enforce contact.
A key observation is that, if we know about self-contact (even approximately), this greatly reduces pose ambiguity by removing degrees of freedom. 
Thus, knowing contact makes the estimation of 3D human pose from 2D images more accurate.
The resulting method,  \smplifyxmc (for SMPLify-X with Mimicked Contact), produces high-quality 3D reference poses and body shapes in correspondence with the images.

Third, to gain even more image variety, we take images from three public datasets \cite{Johnson10,Johnson11,liuLQWTcvpr16DeepFashion} and have them labeled with discrete body-part contacts. 
This results in the {\em Discrete Self-Contact (DSC)} dataset.
To enable this, we define a partitioning of the body into regions that can be in contact.
Given labeled discrete contacts, we extend SMPLify to optimize body shape using image features and the discrete contact labels. 
We call this method \smplifyxdc, for SMPLify with Discrete Self-Contact.

Given the MTP and DSC datasets, we finetune a recent HPS regression network, SPIN \cite{kolotouros2019learning}.
When we have 3D reference poses, \ie~for MTP images, we use these as though they were ground truth and do not optimize them in SPIN. 
When discrete contact annotations are available, \ie~for DSC images, we use \smplifyxdc to optimize the fit in the SPIN training loop.
Fine-tuning SPIN on MTP and DSC significantly improves accuracy of the regressed poses when there is contact (evaluated on 3DPW \cite{vonMarcard2018}).
Surprisingly, the results on non-self-contact poses also improve, suggesting that (1) gathering accurate 3D poses for in-the-wild images is beneficial, and (2) that self-contact can provide valuable constraints that simplify pose estimation.

We call our regression method {\em TUCH} (Towards Understanding Contact in Humans).
Figure \ref{fig:rotateview} illustrates the effect of exploiting self-contact in 3D HPS estimation. 
By training with self-contact, TUCH significantly improves the physical plausibility.

In summary, the key contributions of this paper are:
(1) We introduce TUCH, the first HPS regressor for self-contact poses, trained end-to-end. 
(2) We create a novel dataset of 3D human meshes with realistic contact (3DCP).
(3) We define a ``Mimic The Pose''  {\mtp} task and a new optimization method to create a novel dataset of in-the-wild images with accurate 3D reference data.
(4) We create a large dataset of images with reference poses that use discrete contact labels.
(5) We show in experiments that taking self-contact information into account improves pose estimation in two ways (data and losses), and in turn achieves state-of-the-art results on 3D pose estimation benchmarks.
(6) The data and code are available for research purposes.

%% file: main_sections/02_RelatedWork.tex
\section{Related Work}

{\bf 3D pose estimation with contact.}
Despite rapid progress in  3D human pose estimation~\cite{Joo2018totalcapture,kanazawa2018end,kolotouros2019learning,mehta2019xnect,SMPL-X:2019,sarafianos20163d,xiang2019monocular}, 
and despite the role that self-contact plays in our daily lives, only a handful of previous works discuss self-contact. 
Information about contact can benefit 3D HPS estimation in many ways, usually by providing additional physical constraints to prevent undesirable solutions such as interpenetration between limbs.

\textit{Body contact.} Lee and Chen~\cite{lee1985determination} approximate the human body as a set of line segments and avoid collisions between the limbs and torso. 
Similar ideas are adopted in~\cite{belagiannis20163Dpami,Ganapathi2012eccv} where line segments are replaced with cylinders. Yin~\etal~\cite{yin2018sampling} build a pose prior to penalize 
deep interpenetration detected by the Open Dynamics Engine~\cite{smith2005open}. 
While efficient, these stickman-like representations are far from realistic. 
Using a full 3D body mesh representation, Pavlakos~\etal~\cite{SMPL-X:2019} take advantage of physical limits and resolve interpenetration of body parts by adding an interpenetration loss. 
When estimating multiple people from an image, Zanfir~\etal~\cite{zanfir2018monocular} use a volume occupancy exclusion loss to prevent penetration.
Still, other work has exploited textual and ordinal descriptions of body pose \cite{pavlakos2018ordinal,PonsMoll_CVPR2014}.
This includes constraints like ``Right hand above the hips''.  
These methods, however, do not consider self-contact.

Most similar to us is the work of Fieraru~\etal~\cite{Fieraru_2020_CVPR}, which utilizes discrete contact annotations between people.
They introduce contact signatures between people based on coarse body parts. This is similar to how we collect the DSC dataset.
Contemporaneous with our work, Fieraru et al.~\cite{Fieraru_2021_AAAI} extend this to self-contact with a 2-stage approach.
They train a network to predict ``self-contact signatures", which are used for optimization-based 3D pose estimation.
In contrast, TUCH is trained end-to-end to regress body pose with contact information.

\textit{World contact.} Multiple methods use the 3D scene to help estimate the human pose. 
Physical constraints can come from the ground plane~\cite{vondrak2012dynamical,zanfir2018monocular}, an object~\cite{PROX:2019,kim2014shape2pose,kjellstrom2010tracking}, or contextual scene information~\cite{gupta2008context,yamamoto2000scene}. 
Li~\etal~\cite{li2019estimating} use a DNN to detect 2D contact points between objects and selected body joints. 
Narasimhaswamy~\etal~\cite{Narasimhaswamy2020nips} categorize hand contacts into self, person-person, and object contacts and aim to detect them from in-the-wild images. Their dataset does not provide reference 3D poses or shape. 

All the above works make a similar observation: human pose estimation is not a stand-alone task; considering additional physical contact constraints improves the results. 
We go beyond prior work by addressing self-contact and showing how training with self-contact data improves pose estimation overall.

{\bf 3D body datasets.}
While there are many datasets of 3D human scans, most of these have people standing in an ``A''  or ``T'' pose to explicitly minimize self-contact \cite{Robinette1999TheCP}.
Even when the body is scanned in varied poses, these poses are  designed to avoid self-contact
\cite{Anguelov2005scape,dfaust:CVPR:2017,bronstein2008numerical,Dyna:SIGGRAPH:2015}.
For example, the FAUST dataset has a few examples of self-contact and the authors identify these as the major cause of error for scan processing methods \cite{Bogo:CVPR:2014}. 
Recently, the AMASS~\cite{AMASS:2019} dataset unifies 15 different optical marker-based motion capture (mocap) datasets within a common 3D body parameterization, offering around 170k meshes with SMPL-H~\cite{MANO:SIGGRAPHASIA:2017} topology. 
Since mocap markers are sparse and often do not cover the hands, such datasets typically do not explicitly capture self-contact.
As illustrated in Table~\ref{tab:contactinexistingdatasets}, none of these datasets explicitly addresses self-contact.

\begin{table}
	\begin{center}
		\begin{tabular}{lcc}
			\hline
			Name       & Meshes &   Meshes with self-contact    \\
			\hline
			3DCP Scan (ours) & 190       &    188  \\ \hline
			3D BodyTex \cite{ahmed2018survey} & 400        &    3     \\
			SCAPE \cite{Anguelov2005scape}     & 70              &    0    \\
			Hasler \etal \cite{hasler2009statistical} & 	520      &    0     \\
			FAUST  \cite{Bogo:CVPR:2014}    & 100/ 400            &     20/ 140   \\
			\hline
		\end{tabular}
	    \caption{Existing 3D  human mesh datasets with the number of poses and the number of contact poses identified by visual inspection. 
	    \threedcpscan is the scan subset of \threedcp (see Section \ref{section:SelfContactDatasets}). 
	    FAUST (train/test) includes scans with self-contact, \ie 20 in the training and 140 in the test set. 
	    However, in FAUST the variety is low as each subject is scanned in the same 10/20 poses, whereas in \threedcpscan each subject does different poses.
	    }
	    \label{tab:contactinexistingdatasets}
    \end{center}
\end{table}

{\bf Pose mimicking.}
Our Mimic-The-Pose dataset uses the idea that people can replicate a pose that they are shown.
Several previous works have explored this idea in different contexts.
Taylor et al.~\cite{taylor2011learning} crowd-source images of people in the same pose by imitation.  
While they do not know the true 3D pose, they are able to train a network to match images of  people in similar poses.
Marinoiu et al.~\cite{marinoiu2013pictorial} motion capture subjects reenacting a 3D pose from a 2D image.
They found that subjects replicated 3D poses with a mean joint error of around 100mm.
This is on par with existing 3D pose regression methods, pointing to people's ability to approximately recreate viewed poses.
Fieraru et al.~\cite{Fieraru_2021_AAAI} 
ask subjects to reproduce contact from an image
in a lab setting.
They manually annotate the contact, whereas our MTP task is done in people's homes and \smplifyxmc is used to automatically optimize the pose and contact.

%% file: main_sections/03_SelfContact.tex
\section{Self-Contact}
\label{subsection:Contact}
An intuitive definition of contact between two meshes, e.g.~a human and an object, is based on intersecting triangles. Self-contact, however, must be formulated to exclude common, but not functional, triangle intersections, e.g.~at the crotch or armpits. 
Intuitively, vertices are in self-contact if they are close in Euclidean distance (near zero) but distant in geodesic distance, i.e.~far away on the body surface.
\begin{definition}
	\label{definition-selfcontact}
	Given a mesh $M$ with vertices $M_V$, we define two vertices $v_{i}$ and $v_{j}$ $\in M_V$ to be in \textit{self-contact}, if \begin{enumerate*}[label=(\roman*)] \item $\left\|  v_i - v_j \right\|  < \euclth$, and \item $\mathit{geo}(v_i, v_j) >\geoth$,\end{enumerate*} where $\euclth$ and $\geoth$ are predefined thresholds and $\mathit{geo}(v_i,v_j)$ denotes the geodesic distance between $v_i$ and $v_j$. We use shape-independent geodesic distances precomputed on the neutral, mean-shaped SMPL and SMPL-X models.  
\end{definition}

Following this definition, we denote the set of vertex pairs in self-contact as $\mathnormal{M}_C := \{(v_i, v_j) | v_i, v_j \in M_V \text{ and } v_i, v_j \text{ satisfy Definition~\ref{definition-selfcontact}}\}$.  $M$ is a \textit{self-contact mesh} when $|\mathnormal{M}_C| > 0$. We further define an operator $\mathcal{U}(\cdot)$ that returns a set of unique vertices in $M_C$, and an operator $f_g(\cdot)$ that takes $v_i$ as input and returns the Euclidean distance to the nearest $v_j$ that is far enough in the geodesic sense. 
Formally, $\mathcal{U}(M_C)=\{v_0, v_1, v_2, \dots, v_n\}$, where $\forall v_i \in \mathcal{U}(M_C)\,\ , \exists v_j \in \mathcal{U}(M_C)$, such that $(v_i,v_j) \in M_C$. $f_{g}(v_i) := \min_{v_j \in \mathnormal{M}_{G}(v_i)} \left\|  v_i - v_j\right\| $, where $\mathnormal{M}_{G}(v_i) := \{v_j | \mathit{geo}(v_i,v_j) > \geoth\}$.

We further cluster self-contact meshes into distinct types.
To that end, we define self-contact signatures $\mathbf {S} \in \{0,1\}^{K \times K}$;
see \cite{Fieraru_2021_AAAI} for a similar definition.
We first segment
the vertices of a mesh into $K$ regions $R_k$, where  $R_k \cap R_l = \emptyset$ for $k \neq l$ and $ \bigcup_{k=1}^{K} R_k = M_V $.
We use fine signatures to cluster self-contact meshes from AMASS (see \suppmat) and rough signatures
(see Fig.~\ref{fig:DFExamples}) for human annotation.
\begin{definition}
\label{definition-signature}
Two regions $R_k$ and $R_l$ are in contact if $\exists (v_i, v_j) \in M_C$, such that $v_i \in R_k$ and $v_j \in R_l$ holds. If $R_k$ and $R_l$ are in contact, $\mathbf{S}_{kl} = \mathbf{S}_{lk} = 1$. $M_\mathbf{S}$ denotes the contact signature for mesh $M$. 
\end{definition}

\begin{figure}
\centerline{	\includegraphics[width=\linewidth]{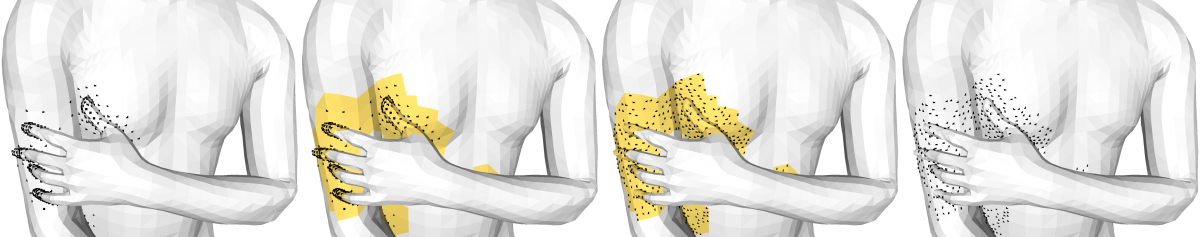}}
	\caption{Visualization of the function $\mathit{HD}(X)$, that maps from mesh vertices to mesh surface points. First, a SMPL-X mesh with vertices in contact highlighted. Second, in yellow, all faces containing a vertex in contact are selected. Then, all points lying on a face containing a vertex in contact are selected from $M_P$, denoted as $M_D$. $M_P$ is a fixed set of mesh surface points that are regressed from mesh vertices. Note that in image one and two the finger vertices are denser than the arm and chest vertices, in contrast to the more uniform density in images three and four.}
	\label{fig:HD_SMPLX}
\end{figure}

To detect self-contact, we need to be able to quickly compute the distance between two points on the body surface.
Vertex-to-vertex distance is a poor approximation of this due to the varying density of vertices across the body.
Consequently, we introduce HD SMPL-X and HD SMPL to efficiently approximate surface-to-surface distance.
For this, we uniformly, and densely, sample points, $M_P \in \mathbb{R}^{P\times3} \text{ with }P=20,000$ on the body. 
A sparse linear regressor $\mathcal{P}$ regresses $M_P$ from the mesh vertices $M_V$, $M_P = \mathcal{P} M_V$. The geodesic distance $geo_{\mathit{HD}}(p_1, p_2)$ 
between $p_1 \in M_P$ and $p_2 \in M_P$ is approximated via $\mathit{geo}(m,n)$, where $m = \arg\min_{v \in M_V} \left\|v - p_1\right\| $ 
and $n = \arg\min_{v \in M_V} \left\| v - p_2\right\| $.
In practice, we use mesh surface points only when contact is present by following a three-step procedure as illustrated in Fig.~\ref{fig:HD_SMPLX}. First, we use 
Definition~\ref{definition-selfcontact} to detect vertices in contact, $M_C$. Then we select all points 
in $M_P$ lying on faces that contain vertices in $M_C$, denoted as $M_D$. Last, for $p_i \in M_D$ we find the 
closest mesh surface point $\min_{p_j \in M_D}\left\| p_i-p_j\right\| $, such that 
$\mathit{geo}_{\mathit{HD}}(p_i, p_j) > \geoth$. With $\mathit{HD}(X): X \subset M_V \rightarrow M_D \subset M_P $ 
we denote the function that maps from a set of mesh vertices to a set of mesh surface points.
As the number of points, $P$, increases, the point-to-point distance approximates the surface-to-surface distance.

%% file: main_sections/04_TheDataset.tex
\section{Self-Contact Datasets}
\label{section:SelfContactDatasets}
Our goal is to create datasets of in-the-wild images paired with 3D human meshes as pseudo ground truth. 
Unlike traditional pipelines that collect images first and then annotate them with pose and shape parameters~\cite{joo2020eft,vonMarcard2018}, we take the opposite approach. 
We first curate meshes with self-contact and then pair them with images through a novel  pose mimicking and fitting procedure. 
We use SMPL-X to create the 3DCP and MTP dataset to better fit contacts between hands and bodies. However, to fine-tune SPIN \cite{kolotouros2019learning}, we convert MTP data to SMPL topology, and use \smplifyxdc when optimizing with discrete contact.

\subsection{3D Contact Pose (3DCP) Meshes}
We create 3D human meshes with self-contact in two ways: with 3D scans and with motion capture data.

\textbf{3DCP Scan.}
We scan 6 subjects (3 males, 3 females) in self-contact poses. 
We then register the SMPL-X mesh topology to the raw scans.
These registrations are obtained using Co-Registration  \cite{Hirshberg:ECCV:2012}, which iteratively deforms the SMPL-X template mesh $V$ to minimize the \emph{point-to-plane} distance between the scan points $S \in \mathbb{R}^{N\times3}$, where $N$ is the number of scan points and the template points $V \in \mathbb{R}^{10375 \times 3}$.  
However, registering poses with self-contact is challenging. When body parts are in close proximity, the standard process can result in interpenetration. 
To address this, we add a self-contact-preserving energy term to the objective function. 
If two vertices $v_{i}$ and $v_{j}$ are in contact according to Definition~\ref{definition-selfcontact}, we minimize the \emph{point-to-plane} distance between triangles including $v_{i}$ and the triangular planes including  $v_{j}$. 
This term ensures that body parts that are in contact remain in contact; see {\suppmat}~for details.

\begin{figure}[t!]
\centerline{
		\includegraphics[width=\linewidth]{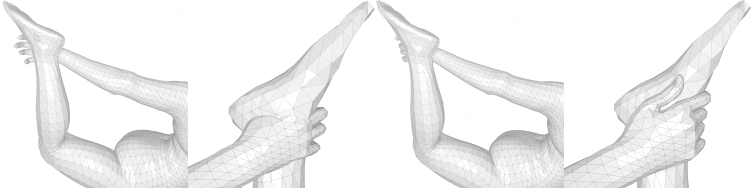}
}
	\caption{Self-contact optimization. Column 1 and 2: a pose selected from AMASS with near self-contact (between the fingertips and the foot) and interpenetration (thumb and foot). Column 3 and 4: after self-contact optimization, all fingers are in contact with the foot and interpenetration is reduced. }
	\label{fig:BeforeAfterSelfContactOptimization}
\end{figure}

\textbf{3DCP Mocap.}
While mocap datasets are usually not explicitly designed to capture self-contact,
it does occur during motion capture. 
We therefore search the AMASS dataset for poses that satisfy our self-contact definition.
We find that some of the selected meshes from AMASS contain small amounts of self-penetration or near contact.
Thus, we perform \emph{self-contact optimization} to fix this while encouraging contact, as shown in Fig.~\ref{fig:BeforeAfterSelfContactOptimization}; see {\suppmat}~for details.

\begin{figure*}[t]
	\includegraphics[width=\textwidth]{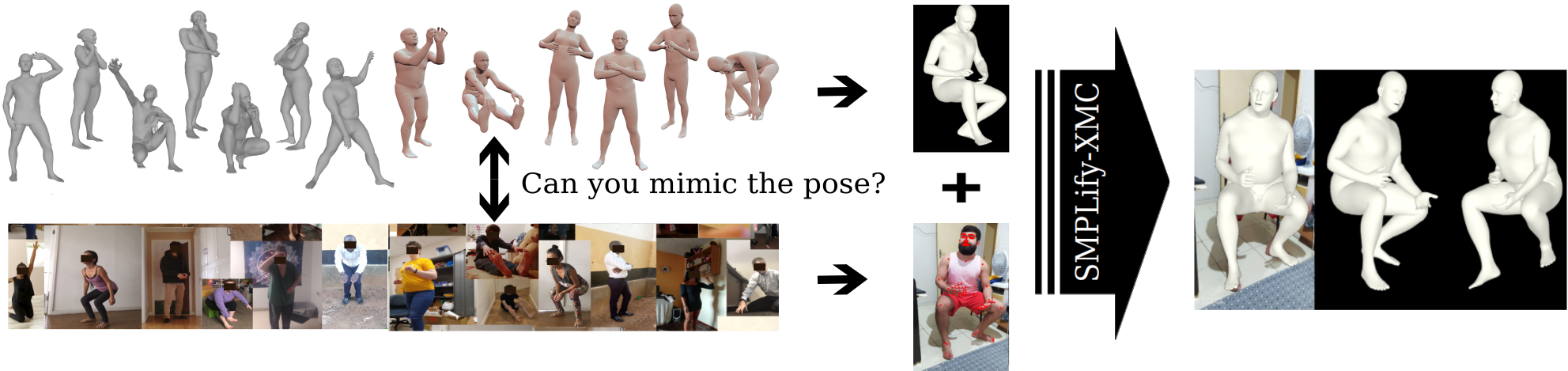}
	\caption{Mimic-The-Pose (MTP) dataset. MTP is built via: (1) collecting many 3D meshes that exhibit self-contact. In grey, new 3D scans in self-contact poses, in brown self-contact poses optimized from AMASS mocap data.  (2) collecting images in the wild, by asking workers on AMT to mimic poses and contacts.  (3) the presented meshes are refined via \smplifyxmc to match the image features.}
	\label{fig:selfcontact-example}
\end{figure*}

\subsection{Mimic-The-Pose (MTP) Data}~\label{subsection-mtp}
\begin{figure*}[t]
\centerline{		\includegraphics[trim=0 6in 0 0, clip=true,width=\textwidth]{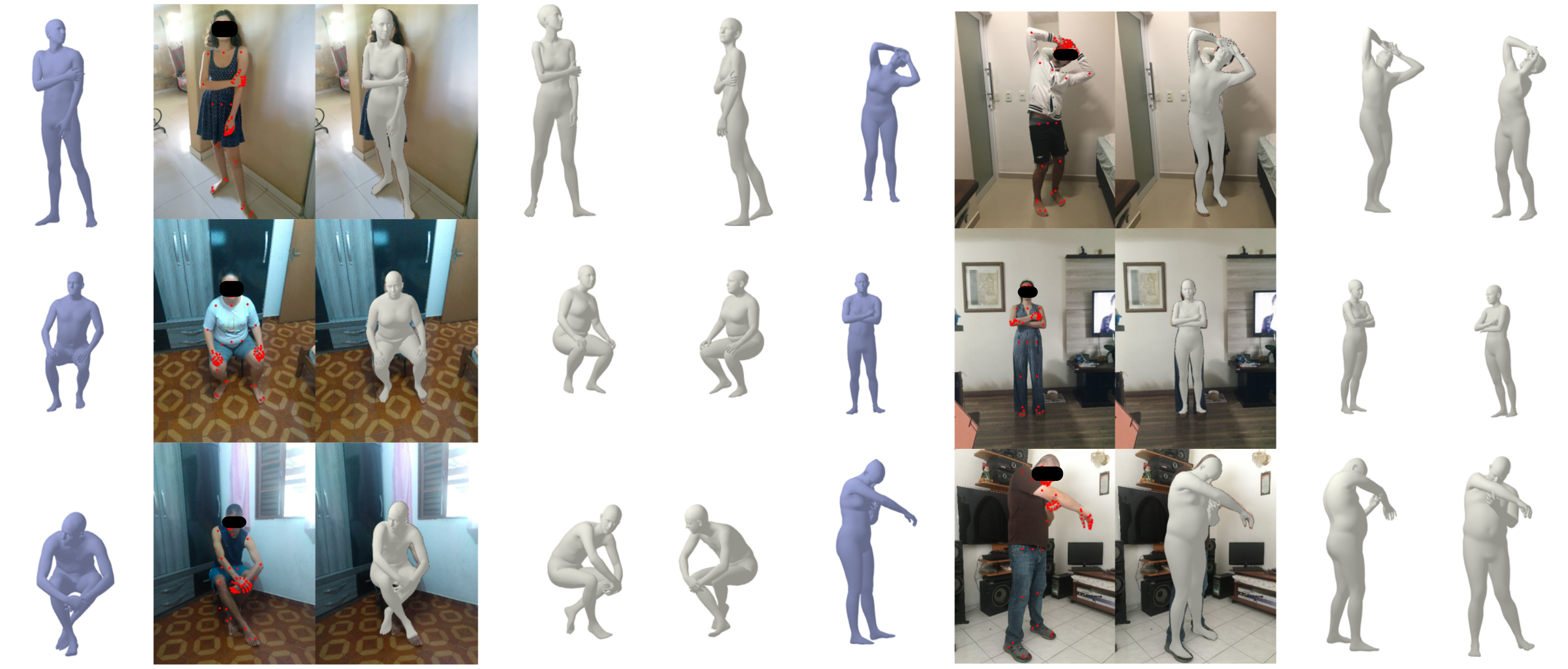}}
\vspace{-0.02in}
	\caption{\mtp results. Meshes presented to AMT workers (blue) and the images they submitted with OpenPose keypoints overlaid. In grey, the pseudo ground-truth meshes computed by \smplifyxmc.}
	\label{fig:smplifyxcfits}
\end{figure*}
To collect in-the-wild images with near ground-truth 3D human meshes,  we propose a novel two-step process (see Fig.~\ref{fig:selfcontact-example}). 
First, using meshes from 3DCP as examples, workers on AMT are asked to mimic the pose as accurately as possible while someone takes their photo showing the full body (the \emph{mimicked pose}).
Mimicking poses may be challenging for people when only a single image of the pose is presented~\cite{marinoiu2013pictorial}. 
Thus, we render each 3DCP mesh from three different views with the contact regions highlighted (the \emph{presented pose}).
We allot 3 hours time for ten poses. Participants also provide their height and weight.
All participants gave informed consent for the capture and the use of their imagery.
Please see {\suppmat}~for details.

\textbf{\smplifyxmc.}
The second step applies a novel optimization method to estimate the pose in the image, given a strong prior from the presented pose.
The presented pose $\tilde{\theta}$, shape $\tilde{\beta}$, and gender is not mimicked perfectly.
To obtain pseudo ground-truth pose and shape, we adapt SMPLify-X~\cite{SMPL-X:2019}, a multi-stage optimization method, that fits  SMPL-X pose $\theta$, shape $\beta$, and expression $\psi$ to image features starting from the mean pose and shape. We make use of the presented pose $\tilde{\theta}$ in three ways: first, to initialize the optimization and solve for global orientation and camera; second, it serves as a pose prior; and third its contact is used to keep relevant body parts close to each other. We refer to this new optimization method as \smplifyxmc.

In the first stage, we optimize body shape $\beta$, camera $\Pi$ (rotation, translations, and focal length), and body global orientation $\theta_g$, while the pose $\theta$ is initialized as $\tilde{\theta}$ and stays constant; see \suppmat~for a description of the first stage.

In the second and third stage, we jointly optimize $\theta$, $\beta$, and $\Pi$  to minimize 
\begin{equation}
\begin{aligned}
\mathcal{L}(\theta, \beta, \Pi) = & E_J + \lambda_{m_h}E_{m_h} + \lambda_{\tilde{\theta}} \mathcal{L}_{\tilde{\theta}} + \\
& \lambda_{M} \mathcal{L}_M + \lambda_{\tilde{C}} \mathcal{L}_{\tilde{C}} + \lambda_{S} \mathcal{L}_{S}.
\end{aligned}
\end{equation}
$E_J$ denotes the same re-projection loss as specified in~\cite{SMPL-X:2019}\footnote{We denote loss terms defined in prior work as $E$ while ours as $\mathcal{L}$.}. We use the standard SMPLify-X  priors for left and right hand $E_{m_h}$. While the pose prior in~\cite{SMPL-X:2019} penalizes deviation from the mean pose, here, $\mathcal{L}_{\tilde{\theta}}$ is an L2-Loss that penalizes deviation from the presented pose. The measurements loss $\mathcal{L}_M$ takes ground-truth height and weight into account; see \suppmat~for details. The term $\mathcal{L}_{\tilde{C}}$ acts on $\tilde{M}_C$, 
the vertices in self-contact on the presented mesh.
To ensure the desired self-contact, one could seek to minimize the distances between vertices in contact, e.g.~$||v_i - v_j||$, $(v_i, v_j) \in \tilde{M}_C$. However, with this approach, we observe slight mesh distortions, when presented and mimicked contact are different.
Instead, we use a term that encourages every vertex in $\tilde{M}_C$ to be in contact. More formally, 
\begin{equation}
\mathcal{L}_{\tilde{C}} = \frac{1}{|\mathcal{U}(\tilde{M}_C)|} \sum_{v_i \in \mathcal{U}(\tilde{M}_C)} \tanh(f_{g}(v_i)).
\end{equation}

The third stage actives $\mathcal{L}_{S}$  for fine-grained self-contact optimization, which resolves interpenetration while encouraging contact. 
The objective is $\mathcal{L_S} = \lambda_{C} \mathcal{L}_{C} + \lambda_{P} \mathcal{L}_{P} +  \lambda_{A} \mathcal{L}_{A}$.
Vertices in contact are pulled together via a contact term $\mathcal{L}_{C}$; vertices inside the mesh are pushed to the surface via a pushing term $\mathcal{L}_{P}$, and $\mathcal{L}_{A}$ aligns the surface normals of two vertices in contact.

To compute these terms, we must first find which vertices are inside, $M_{I} \subset M_V$, or in contact, $M_{C} \subset M_V$. 
$M_{C}$ is computed following Definition~\ref{definition-selfcontact} with $\geoth = 30$cm and $\euclth = 2$cm.
The set of inside vertices $M_{I}$ is detected by generalized winding numbers~\cite{jacobson2013robust}. 
SMPL-X is not a closed mesh and thus complicating the test for penetration.
Consequently, we close it by adding a vertex at the back of the mouth.
In addition, neighboring parts of SMPL and SMPL-X often intersect, e.g.~torso and upper arms.
We identify such common self-intersections and filter them out from $M_{I}$. See Sup.~Mat.~for details.
To capture fine-grained contact, we map the union of inside and contact vertices onto the HD SMPL-X surface, \ie~$M_D = \mathit{HD}(M_{I} \cup M_{C})$, 
which is further segmented into an inside $M_{D_I}$ and outside $M_{D_I^{\complement}}$ subsets by testing for intersections. 
The self-contact objectives are defined as 
\begin{align*}
\mathcal{L}_{C} = & \sum_{p_i \in M_{D_I^{\complement}}} \alpha_1 \tanh(\frac{f_{g}(p_i)}{\alpha_2})^2 \text{,} \\
\mathcal{L}_{P} = & \sum_{p_i \in M_{D_I}} \beta_1 \tanh(\frac{f_{g}(p_i)}{\beta_2})^2 \text{,} \\
\mathcal{L}_{A} = & \sum_{(p_i, p_j) \in M_{D_C}} 1 + \inp{N(p_i)}{N(p_j)} \text{.}
\end{align*}
$f_{g}$ denotes the function that finds the closest point $p_j \in M_D$. $M_{D_C}$ is the subset of vertices in contact in $M_D$. 
We use $\alpha_1 = \alpha_2 = 0.005$, $\beta_1 = 1.0$, and $\beta_2 = 0.04$ and visualize the contact and pushing functions in the {\suppmat} 
Fig.~\ref{fig:smplifyxcfits} shows examples of our pseudo ground-truth meshes.

\subsection{Discrete Self-Contact (DSC) Data}~\label{subsection-dsc}
\begin{figure}
\centerline{		\includegraphics[width=0.95\linewidth]{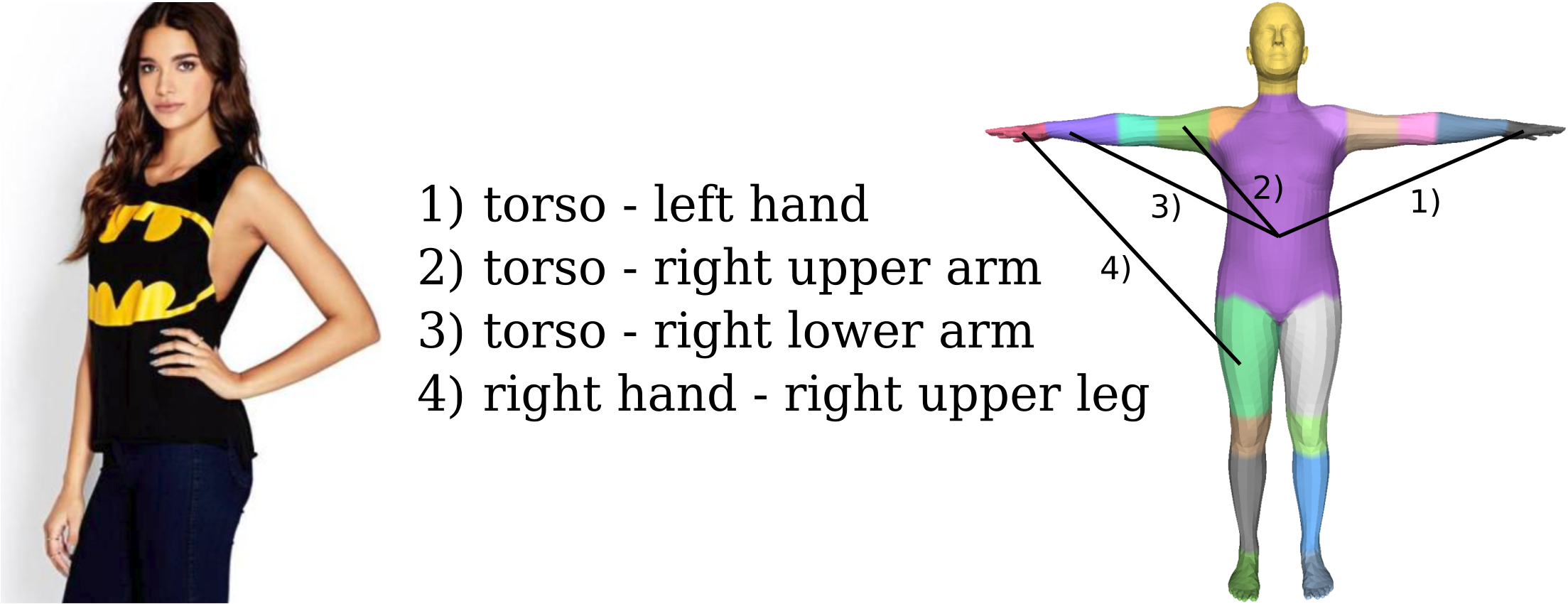} }
\vspace{-0.04in}
	\caption{DSC dataset. Image with discrete contact annotation on the left. Right: DSC signature with $K=24$ regions.}
	\label{fig:DFExamples}
\end{figure}

Images in the wild collected for human pose estimation normally come with 2D keypoint annotations, body segmentation, or bounding boxes. 
Such annotations lack 3D information. 
Discrete self-contact annotation, however, provides useful 3D information about pose. 
We use $K = 24$ regions and label their pairwise contact for three publicly available datasets, namely Leeds Sports Pose (LSP), Leeds Sports Pose Extended (LSPet), and DeepFashion (DF). An example annotation is visualized in Fig.~\ref{fig:DFExamples}.
Of course, such labels are noisy because it can be difficult to accurately determine contact from an image. See \suppmat~for details.

\subsection{Summary of the Collected Data}
Our 3DCP human mesh dataset consists of 190 meshes containing self-contact from 6 subjects, 159 SMPL-X bodies fit to commercial scans from AGORA \cite{patel2021agora}, and 1304 self-contact optimized meshes from mocap data. From these 1653 poses, we collect 3731 mimicked pose images from 148 unique subjects (52 female; 96 male) for MTP and fit pseudo ground-truth SMPL-X parameters. MTP is diverse in body shapes and ethnicities. Our DSC dataset provides annotations for 30K images.

%% file: main_sections/05_Methods.tex
\section{TUCH}
\label{sec:tuch}
Finally, we train a regression network that has the same design as SPIN~\cite{kolotouros2019learning}. At each training iteration, the current regressor estimates the pose, shape, and camera parameters of the SMPL model for an input image. 
Using ground-truth 2D keypoints, an optimizer refines the estimated pose and shape, which are used, in turn, to supervise the  regressor.
We follow this regression-optimization scheme for \dsc data, where we have no 3D ground truth. 
To this end, we adapt the in-the-loop SMPLify routine to account for discrete self-contact labels, which we term \smplifyxdc.
For \mtp images, we use the pseudo ground truth from \smplifyxmc as direct supervision with no optimization involved.
We explain the losses of each routine below.

\textbf{Regressor.} Similar to SPIN, the regressor of TUCH predicts pose, shape, and camera, with the loss function:
\begin{equation}
L_R = E_J + \lambda_{\theta} E_{\theta} + \lambda_{\beta} E_{\beta} + \lambda_{C} \mathcal{L}_{C} + \lambda_{P} \mathcal{L}_{P} \text{.}
\end{equation}
$E_J$ denotes the joint re-projection loss. $\mathcal{L}_{P}$ and $\mathcal{L}_{C}$ are self-contact loss terms used in $\mathcal{L}_{S}$ in \smplifyxmc, where $\mathcal{L}_{P}$ penalizes mesh intersections and $\mathcal{L}_{C}$ encourages contact. Further, $E_{\theta}$ and $E_{\beta}$ are L2-Losses that penalize deviation from the pseudo ground-truth pose and shape.

\textbf{Optimizer.} We develop \smplifyxdc to fit pose $\theta_{opt}$, shape $\beta_{opt}$, and camera $\Pi_{opt}$ to DSC data, taking ground-truth keypoints and contact as constraints.
\begin{figure}[t]
	\centerline{\includegraphics[width=\linewidth]{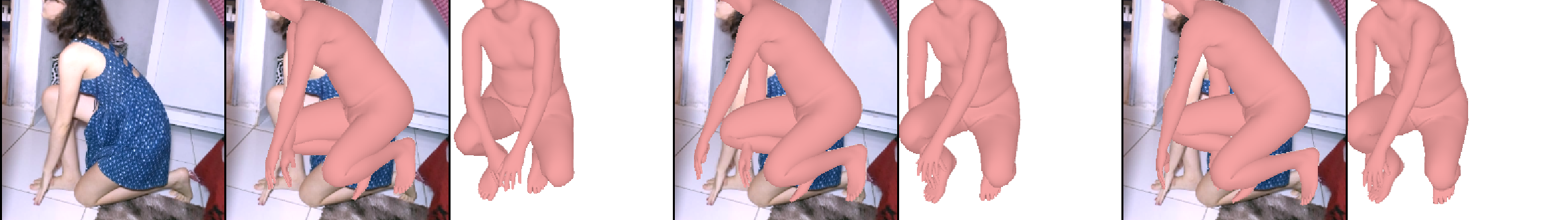}}
	\caption{Initial wrong contact (left) from the regressor is fixed by \smplifyxdc after 5 (middle) and 10 (right) iterations.}
	\label{fig:Optimization_Wrong_Contact}
\end{figure}
Typically, in human mesh optimization methods the camera is fit first, then the model parameters follow. However, we find that this can distort  body shape when encouraging contact. Therefore, we optimize shape and camera translation first, using the same camera fitting loss as in \cite{kolotouros2019learning}. After that, body pose and global orientation are optimized under the objective 
\begin{equation}
L_O(\theta) = E_J + \lambda_{\theta}E_{\theta} + \lambda_{C} \mathcal{L}_{C} + \lambda_{P} \mathcal{L}_{P} + \lambda_{D} \mathcal{L_D} \text{.}
\label{eq:DSCOptimization}
\end{equation}
The discrete contact loss, $\mathcal{L_D}$, penalizes the minimum distance between regions in contact. 
Formally, given a contact signature $\mathbf{S}$ where $\mathbf{S}_{ij} = \mathbf{S}_{ji} = 1$ if two regions $R_i$ and $R_j$ are annotated to be in contact, we define $$\mathcal{L_D} =\sum_{i = 1}^{K} \sum_{j = i+1}^{K} \mathbf{S}_{ij} \min_{v \in R_i, u \in R_j} ||v - u||^2 \text{.}$$ 
Given the optimized pose $\theta_{opt}$, shape $\beta_{opt}$, and camera $\Pi_{opt}$, we compute the re-projection error and the minimum distance between the regions in contact.
When the re-projection error improves, and more regions with contact annotations are closer than before, we keep the optimized pose as the current best fit. 
When no ground truth is available, the current best fits are used to train the regressor. 

\begin{figure*}[t]
\centerline{		\includegraphics[width=\linewidth]{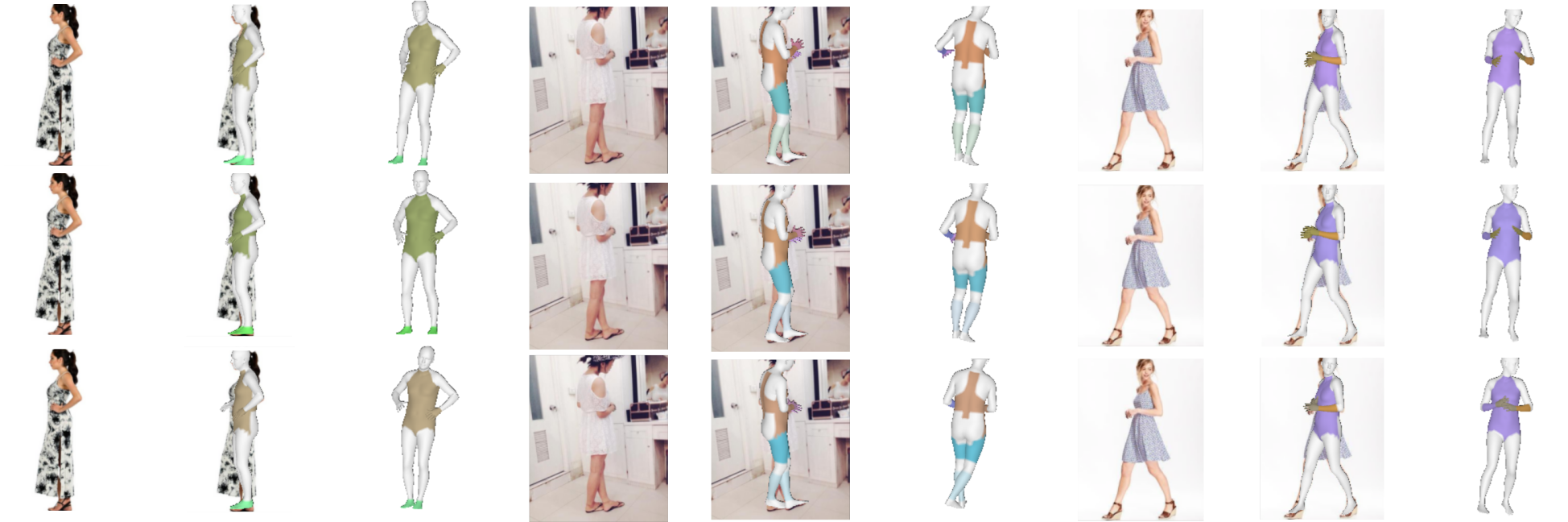}}
\vspace{-0.01in}
	\caption{Impact of discrete self-contact labels in human pose estimation. Body parts labeled in contact are shown in the same color. First row shows an initial SPIN estimate, second row the SMPLify fit, third row the \smplifyxdc fit after 20 iterations.}
	\label{fig:DSCOptimization}
\end{figure*}

We make three observations: (1) The optimizer is often able to fix incorrect poses estimated by the regressor because it considers the ground-truth keypoints and contact (see Fig.~\ref{fig:Optimization_Wrong_Contact}).
(2) Discrete contact labels bring overall improvement by helping resolve depth ambiguity (see Fig.~\ref{fig:DSCOptimization}).
(3) Since we have mixed data in each mini-batch, the direct supervision of MTP data 
improves the regressor, which benefits \smplifyxdc by providing better initial estimates.

\textbf{Implementation details.} We initialize our regression network with SPIN weights \cite{kolotouros2019learning}. For \smplifyxdc, we run 10 iterations per stage and do not use the HD operator to speed up the optimization process. 
For the 2D re-projection loss, we use ground-truth keypoints when available and, for MTP and DF images, OpenPose detections weighted by confidence.
From DSC data we only use images where the full body is visible and ignore 
annotated region pairs that are connected in the DSC segmentation (see \suppmat). 

%% file: main_sections/06_Evaluation.tex
\begin{table}[]
	\centering
	\resizebox{0.65\columnwidth}{!}{
		\setlength\tabcolsep{2pt}
		\setlength{\extrarowheight}{2pt}
		\begin{tabular}{lcccc}
			\toprule[1pt]
			& \multicolumn{2}{c}{MPJPE}     & \multicolumn{2}{c}{PA-MPJPE}  \\  \cline{2-5} 
			& 3DPW & MI        & 3DPW & MI      \\   
			\hline
			SPIN \cite{kolotouros2019learning}       & 96.9 & 105.2          & 59.2 & \textbf{67.5}          \\
			EFT \cite{joo2020eft}       & - & -     & \textbf{54.2} & 68.0          \\
			TUCH & \textbf{84.9} & \textbf{101.2}     & 55.5 & 68.6          \\
			\bottomrule[1pt]
		\end{tabular}
	}
	\caption{Evaluation on 3DPW and MPI-INF-3DHP (MI). Bold numbers indicate the best result; units are \emph{mm}. We report the EFT result denoted in their publication when 3DPW was not part of the training data. Please note that SPIN is trained on MI, but we do not include MI in the fine-tuning set. MI contains mostly indoor lab sequences (100\% train, 75\% test), while DSC and MTP contain only in-the-wild images. This domain gap likely explains the decreased performance in PA-MPJPE.}
	\label{tab:results_wo3dpw}
\end{table}

\section{Evaluation}
\begin{table}[t]
	\centering
	\resizebox{0.65\columnwidth}{!}{
		\setlength\tabcolsep{2pt}
		\setlength{\extrarowheight}{2pt}
		\begin{tabular}{lccc}
			\toprule[1pt]
		               & MPJPE & PA-MPJPE & MV2VE \\
		    \hline
                 SPIN \cite{kolotouros2019learning}   & 79.7           & 50.6              & 95.7  \\
			     EFT \cite{joo2020eft}   & 71.4           & 48.3              & 83.9  \\
			     TUCH   & \textbf{69.5}  & \textbf{42.5}     & \textbf{81.5}       \\
			\bottomrule[1pt]  
		\end{tabular}
}
	\caption{Evaluation on \threedcpscan.
	Numbers are in \emph{mm}. Note that in contrast to TUCH, this version of SPIN did not see poses in the MTP dataset during training. Please see Table \ref{tab:ablation_study} and the corresponding text for an ablation study.}
	\label{tab:threedcpscanResult}
\end{table}
We  evaluate TUCH  on the following three datasets:
\textbf{3DPW}~\cite{vonMarcard2018}, \textbf{MPI-INF-3DHP}~\cite{mono-3dhp2017},
and \textbf{\threedcpscan}. 
This latter dataset consists of RGB images taken during the \threedcpscan scanning process.
While TUCH has never seen these images or subjects, the contact poses were mimicked in creation of MTP, which is used in training.

We use standard evaluation metrics for 3D pose, namely Mean Per-Joint
Position Error (MPJPE) and the Procrustes-aligned version (PA-MPJPE),
and Mean Vertex-to-Vertex Error (MV2VE) for shape and contact. 
Tables~\ref{tab:results_wo3dpw} and \ref{tab:threedcpscanResult} summarize the results of TUCH on \threedpw and \threedcpscan.
Interestingly, TUCH is more accurate than SPIN on 3DPW.
See \suppmat~for results of fine-tuning EFT.

We further evaluate our results w.r.t.~contact. 
To this end, we divide the 3DPW test set into subsets, namely for
$t_{geo} = 50$cm: \textit{self-contact} ($t_{eucl} < 1$cm), \textit{no
  self-contact} ($t_{eucl} > 5$cm), and \textit{unclear} ($ 1$cm $ <
t_{eucl} < 5$cm). For 3DPW we obtain 8752 \textit{self-contact}, 16752
\textit{no self-contact}, and 9491 \textit{unclear} poses. 
Table~\ref{tab:results_3DPW_split} shows a clear improvement on poses
with contact and unclear poses compared to a smaller improvement on
poses without contact.

\begin{figure}
\centerline{	
	\includegraphics[width=0.95\linewidth]{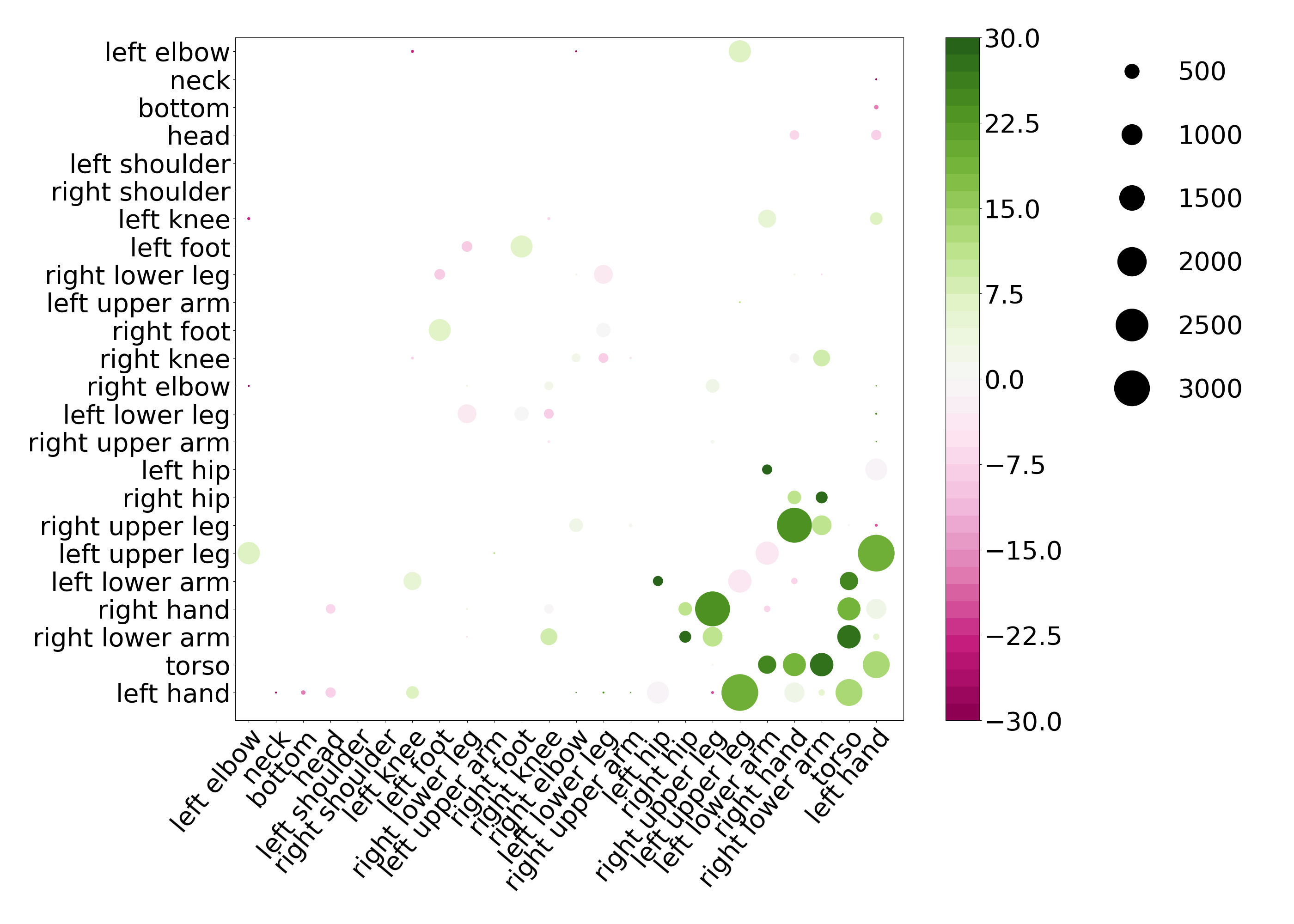}}
	\caption{Average MPJPE difference (SPIN - TUCH), evaluated on the \textit{self-contact} subset of 3DPW. The axes show labels for the  DSC regions. Green indicates that TUCH has a lower error than SPIN on average across all poses with the corresponding regions in contact. The circle size represents the number of images per region. Regions with small circle sizes are less common.}
	\label{fig:MPJPE_Diff}
\end{figure}

\begin{table}[t]
	\centering
	\resizebox{\columnwidth}{!}{
	\setlength\tabcolsep{1.5pt}
	\setlength{\extrarowheight}{2pt}
		\begin{tabular}{lcccc|cccc}
			\toprule[1pt]
	     	 &  \multicolumn{4}{c}{MPJPE}                     & \multicolumn{4}{c}{PA-MPJPE}                  \\ \cline{2-9} 
			& contact          & no contact       & unclear     & \underline{total}  & contact          & no contact       & unclear    & \underline{total}   \\ \hline
			SPIN      & 100.2         & 95.5          & 96.7    & 96.9      & 59.1          & 61.7          & 55.7  &    59.2    \\
			TUCH      & \textbf{85.1} & \textbf{86.6} & \textbf{81.9} & \textbf{84.9} & \textbf{54.1} & \textbf{58.6} & \textbf{51.2} & \textbf{55.5} \\ 
			\bottomrule[1pt]
		\end{tabular}
}
	\caption{Evaluation of TUCH for contact classes in 3DPW. Numbers are in \emph{mm}. See text.
}
	\label{tab:results_3DPW_split}
\end{table}
To further understand the improvement of TUCH over SPIN, we break down the improved MPJPE in 3DPW \textit{self-contact} into the pairwise body-part contact labels defined in the DSC dataset.
Specifically, for each contact pair, we search all poses in 3DPW \textit{self-contact} that have this particular self-contact. We find a clear improvement for a large number of contacts between two body parts, frequently between arms and torso, or e.g.~left hand and right elbow, which is common in arms-crossed poses (see Fig.~\ref{fig:MPJPE_Diff}).

TUCH incorporates self-contact in various ways: annotations of training data, in-the-loop fitting, and in the regression loss. 
We evaluate the impact of each in Table \ref{tab:ablation_study}. 
S+ is SPIN but it sees MTP+DSC images in fine-tuning and runs standard in-the-loop SMPLify with no contact information.
S++ is S+ but uses pseudo ground truth computed with \smplifyxmc on MTP
images; thus self-contact is used to generate the data but nowhere else.
S+ vs.~SPIN suggests that, while poses in \threedcpscan appear in MTP,
just seeing similar poses for training and testing does not yield improvement.
S+ vs.~TUCH is a fair comparison as both see the same images during training. The improved results of TUCH confirm the benefit of using self-contact.
\begin{table}[t]
	\centering
	\begin{adjustbox}{max width=\linewidth}
		\begin{tabular}{lcccc}
		\toprule[1pt]
			& SPIN         & S+        & S++       & TUCH         \\ \cline{1-5} 
			3DPW    & 96.9/ 59.2  & 96.1/ 61.4  & 85.0/ 56.3  & \textbf{84.9}/ \textbf{55.5}  \\
			\threedcpscan & 82.2/ 52.1  & 86.9/ 52.3  & \textbf{74.8}/ 45.7  & 75.2/ \textbf{45.4}  \\
			MI      & 105.2/ 67.5 & 105.8/ 69.4 & 103.1/ 69.0 & \textbf{101.2}/ \textbf{68.6} \\
		\bottomrule[1pt]
		\end{tabular}
	\end{adjustbox}
	\caption{MPJPE/PA-MPJPE (mm) to examine the impact of data and algorithm on 3DPW, \threedcpscan, and MPI-INF-3DHP (MI).}
	\label{tab:ablation_study}
\end{table}

%% file: main_sections/07_Conclusion.tex
\section{Conclusion}
In this work, we address the problem of HPS estimation when self-contact is present. Self-contact is a natural, common occurrence in 
everyday life, but SOTA methods fail to estimate it. One reason for this is that no datasets pairing images in the wild and 3D reference poses exist. 
To address this problem we introduce a new way of collecting data: we ask humans to mimic presented 3D poses. Then we use our new \smplifyxmc method to
fit pseudo ground-truth 3D meshes to the mimicked images, using the presented pose and self-contact to constrain the optimization. We use the new MTP data
along with discrete self-contact annotations to train TUCH; the first end-to-end HPS regressor that also handles poses with self-contact. TUCH uses MTP data 
as if it was ground truth, while the discrete, DSC, data is exploited  during SPIN training via \smplifyxdc. Overall, incorporating contact improves accuracy 
on standard benchmarks like 3DPW, remarkably, not only for poses with self-contact, but also for poses without self-contact. 

%% file: supp_sections/supp_04_TheDataset.tex
\begin{figure*}[h!]
	\includegraphics[width=\linewidth]{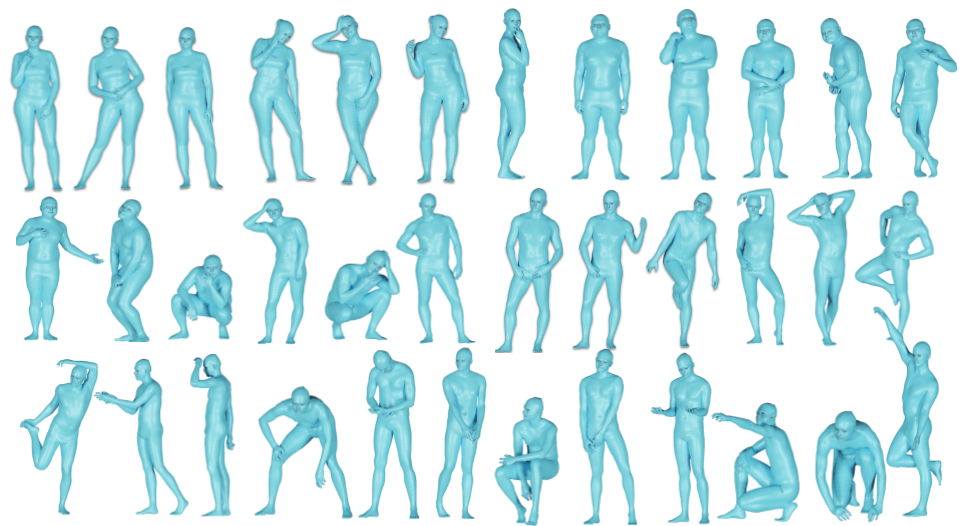}
	\caption{A representative sample from the registrations. A total of 3 male and 3 female subjects were scanned in a diversity of poses that involve self-contact. The 3D scans are registered to a common mesh topology by fitting the SMPL-X template mesh to them using a self-contact preserving energy term that penalizes body part interpenetration.}
	\label{fig:3dcp}
\end{figure*}
The Supplementary Material provides additional details about our methods and visualizations of results.
\section{Self-Contact Datasets}
\label{sec-registration}
\subsection{3D Contact Pose (3DCP) Meshes}
\subsubsection{3DCP Scan} \label{subsec-3dcpscan}
Raw scans have varying topology.  To bring a corpus of scans to a common topology is the process of ``registration''.
Most traditional registration methods ignore interpenetration and self-contact.
Registering our self-contact scans without modeling self-contact would result in self-penetration, particularly where the extremities contact the body.
We address this by modifying the registrations objective function to encourage self-contact without penetration.

Specifically, the fitting objective includes a data term $E_S$ evaluating the goodness of fit of the the vertices $x_v$ on the template  $V$ to  $n$ randomly sampled points, $x_s$ on the surface of the scan $S$ 
\begin{equation}
\label{eq:co_registrations}
E_S( S;V) = \frac{1}{n} \int_{x_s \in \mathcal{S}} \rho \left( \left\lVert  x_s - x_v  \right\lVert \right)
\end{equation}
where $\rho$ is the  Geman-McClure robust penalty function. 

Additionally, we introduce a  self-contact preserving energy term $E_C$ to the objective function. The term $E_C$ helps  to minimize and preserve the \emph{point-to-plane} distance between body parts that are in contact. $E_C$ considers the set of contacting vertex pairs $M_C$ defined by Definition 3.1 
in the main paper. For each tuple $(v_i,v_j)$ in $M_C$, we minimize the \emph{point-to-plane} distance between triangles including $v_{i}$ and the triangular planes including  $v_{j}$.  The contact energy term ensures that body parts that are in contact remain in contact. 

The objective function is minimized in two steps: first a model fitting step, where it is minimized with respect to the SMPL-X model pose parameters $\vec{\theta} \in \mathbb{R}^{55\times3}$  and body shape parameters $\vec{\beta}\in\mathbb{R}^{25}$. Following model fitting, a model-free optimization step minimizes  point-to-plane distance between the model vertices $x_v$ and the scan. 
A sample of the  registrations is shown in Figure~\ref{fig:3dcp}.
\subsubsection{3DCP Mocap.} 
\begin{figure}[t]
	\centerline{	\includegraphics[width=\linewidth]{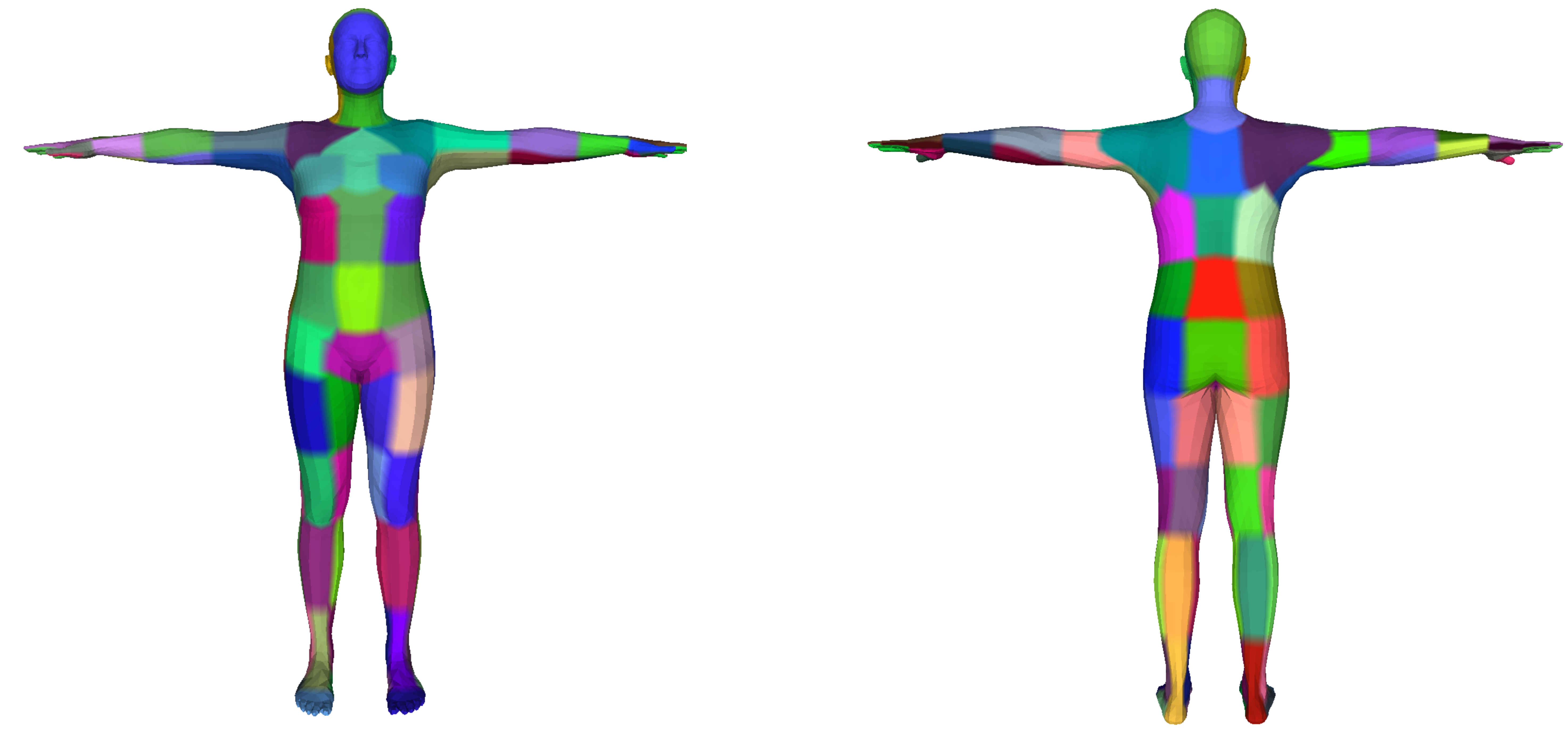}
	}
	\caption{To compute self-contact signatures, we group vertices into distinct regions, shown here with different colors.
		This is useful for searching our scan datasets for poses with specific types of contact.}
	\label{fig:csigSMPLSMPLX}
\end{figure}

\begin{figure}[t!]
\centerline{
\includegraphics[width=\linewidth]{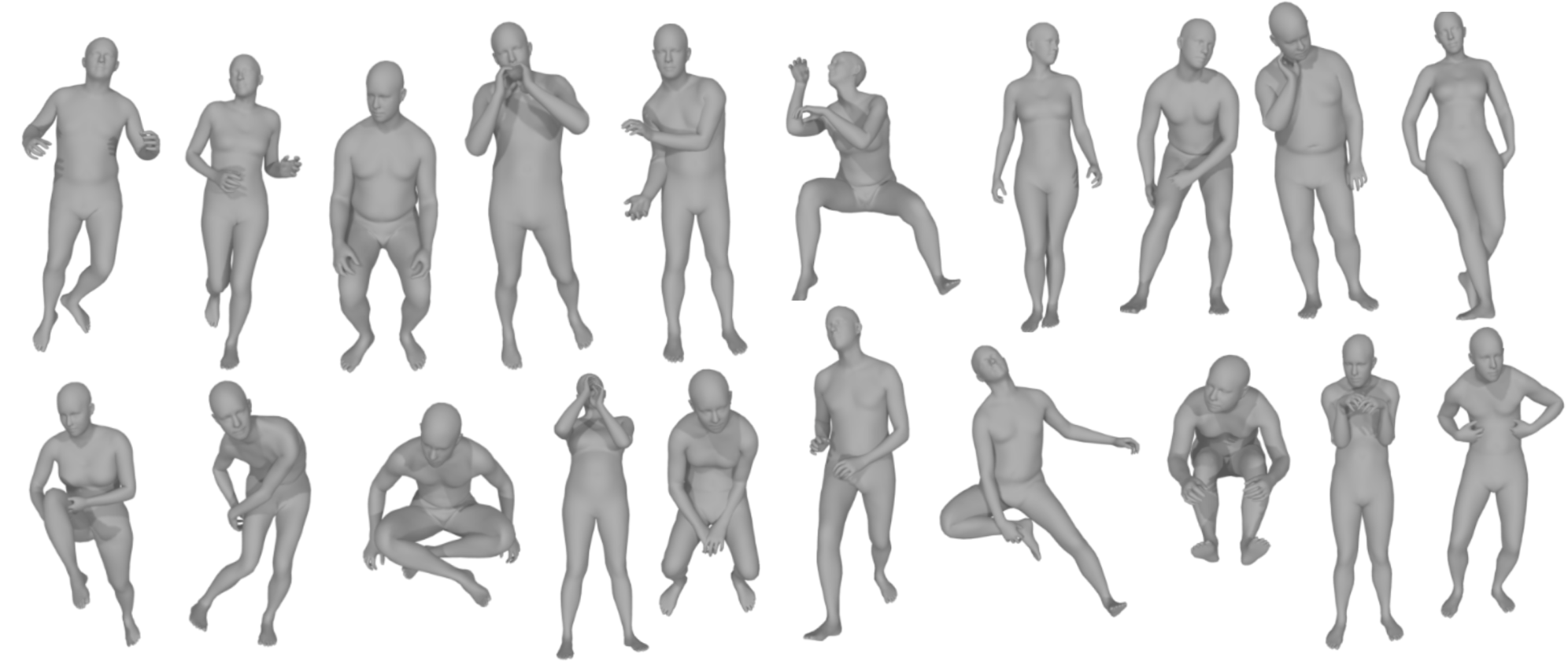}}
\vspace{-0.06in}
	\caption{Sample poses from 20 unique contact signatures, $\mathbf{S}$. We apply $\mathbf {S}$ to select interesting self-contact poses in AMASS.}
	\label{fig:RandomSelectionInterestingSelfContactPoses1Perc}
\end{figure}

\textbf{Sampling meshes from AMASS.} 
First, each mocap.~sequence is sampled at half of its original frame rate. 
For each sampled mesh, we compute the contact signatures $M_\mathbf {S}$ with $\euclth = 3$cm, $\geoth = 30$cm and $K=98$.
The regions are visualized in Fig.~\ref{fig:csigSMPLSMPLX}.
We select only one pose for each unique signature, while ignoring contact when it occurs in more than 1\% of the data. We obtain a subset of {\nummeshes} poses with unique self-contact signatures, as shown in Fig.~\ref{fig:RandomSelectionInterestingSelfContactPoses1Perc}. 

\textbf{Self-Contact Optimization.} Here we provide details of the self-contact optimization for body meshes from the AMASS dataset. In this optimization, vertex pairs in $M_C$ are further pulled together via a contact term $\mathcal{L}_{C}$ and vertices inside the mesh are pushed to the surface via a pushing term $\mathcal{L}_{P}$, while $\mathcal{L}_{O}$ ensures that vertices far away from contact regions stay in place. Note that $\mathcal{L}_{P}$ and $\mathcal{L}_{C}$ are slightly different from the loss terms in the main paper. $\mathcal{L}_{H}$ is a prior for contact between hand and body and $\mathcal{L}_{A}$ aligns the vertex normals when contact happens.

Given the set of vertices $M_V$ of mesh $M$, $M_{E} \subset M_V$ denotes the subset of vertices affiliated with extremities, $M_{I} \subset M_V$ denotes the subset of vertices inside the mesh, and $M_{EI} = M_{E} \cap M_{I}$ denotes the vertices of extremities that are inside the mesh itself and $M_{EI}^\complement$ its complement. 
We identify vertices inside the mesh using generalized winding numbers~\cite{jacobson2013robust}. $M_{V_H} \subset M_V$ is the subset of hand vertices. Note that we make SMPL-X watertight by closing the back of the mouth. $M_{C}$ is computed following Definition 3.1 
in the main paper with $\geoth = 30$cm and $\euclth = 3$cm and $\mathnormal{M}_{G}(v_i) = \{v_j | geo(v_i,v_j) > \geoth \}$. Given an initial mesh $\tilde{I}$, we aim to minimize the objective function

\begin{align}
\label{eq-selfcontactoptimization}
\mathcal{L}(\theta_b, \theta_{h_l}, \theta_{h_r}) = & \lambda_{C} \mathcal{L}_{C} + \lambda_{P} \mathcal{L}_{P} +  \lambda_{H} \mathcal{L}_{H} + \nonumber \\
& \lambda_{O} \mathcal{L}_{O} + \lambda_{A} \mathcal{L}_{A} + \lambda_{\theta_h} \mathcal{L}_{\theta_h} \\
& \lambda_{\theta} \mathcal{L}_{\theta},
\end{align}
where $\theta_{h}$ denote the hand pose vector of the SMPL-X model. Further, 

$$ \mathcal{L}_{C} = \frac{1}{|M_{EI}^\complement|} \sum_{v_i \in M_{EI}^\complement} a \alpha 
 \tanh(\frac{f_{g}(v_i)}{\alpha}),$$

$$ \mathcal{L}_{P} = \frac{1}{|M_{EI}|} \sum_{v_i \in M_{EI}} \gamma_1 \tanh(\frac{f_{g}(v_i)}{\gamma_2}) \text{, and}$$

$$ \mathcal{L}_{H} = \frac{1}{|M_{V_H}|} \sum_{v_i \in M_{V_H}} \delta_1 h_{v_i} \tanh(\frac{f_{g}(v_i)}{\delta_2}),$$
where $f_{g}$ denotes a function, that for each vertex $v_i$ finds the closest vertex in self contact $v_j$, or mathematically $f_{g}(v_i) = \min_{v_j \in \mathnormal{M}_{G}(v_i)} || v_i - v_j||_2$. $h_{v_i}$ denotes the weight per hand vertex from the hand-on-body prior $\mathcal{L}_H$ as explained below, if $v_i$ is outside, otherwise $h_{v_i}=1$. Further, $a= (\min_{v_j \in \mathcal{U}(M_C)} geo(v_i,v_j) + 1)^{-1}$ is an attraction weight. This weight is higher, for vertices close to vertices in contact of $\tilde{I}$. $\mathcal{L}_{\theta}$ is a $L_2$ prior that penalizes deviation from the initial pose and $\mathcal{L}_{\theta_h}$ defines an  $L_2$ prior on the left and right hand pose using the a low-dimensional hand pose space.
$\alpha = 0.04$, $\gamma_1 = 0.07$, $\gamma_2=0.06$ define slope and offset of the pulling and pushing terms. For the hand-on-body-prior we use $\delta_1 = 0.023$, and $\delta_2 = 0.02$ if $v_i$ is inside and $\delta_1 = \delta_2 = 0.01$ if $v_i$ is outside the mesh.

Self-contact optimization aims to correct interpenetration and encourage near-contact vertices to be in contact by slightly refining the poses around the contact regions. 
Vertices that are not affected should stay as close to the original positions as possible.
In $\mathcal{L}_{O}$, the displacement of each vertex from its initial position is weighted by its geodesic distance to a vertex in contact. 
Given $\tilde{v}_i$ denoting the position of vertex $i$ of $\tilde{I}$, the outside loss term is
$$\mathcal{L}_{O} = \delta_2 \sum_{v_i \in M_V}  \min_{v_j \in \mathcal{U}(M_C)} geo(v_i,v_j)^2 ||v_i - \tilde{v}_i||_2,$$
where $\min_{v_j \in \mathcal{U}(M_C)} geo(v_i,v_j) = 1$ if $M_C = \emptyset$ and $\delta_2 = 4$. 
Lastly, we use a term, $\mathcal{L}_{A}$, that encourages the vertex normals $N(v)$ of vertices in contact to be aligned but in opposite directions:
$$ \mathcal{L}_{A} = \frac{1}{|M_{C}|} \sum_{(v_i, v_j) \in M_C} 1 + \inp{N(v_i)}{N(v_j)}.$$

\begin{figure}[t]
	\begin{center}
		\includegraphics[width=\linewidth]{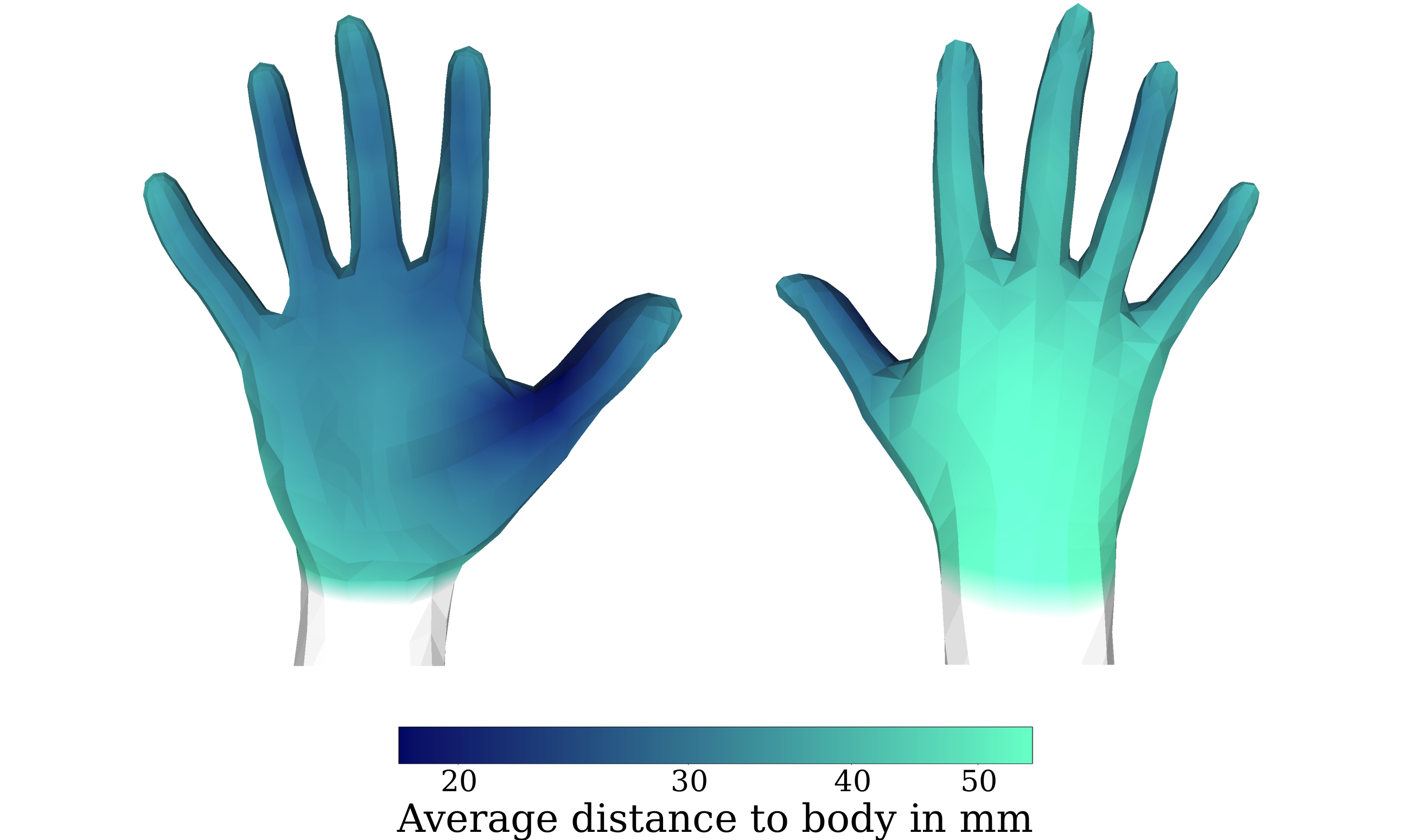}
	\end{center}
	\caption{Hand on body prior. Dark blue indicates small distances to body on average across all registrations where hands are close to the body. The prior is identical for left and right hand.}
	\label{fig:handonbodyprior}
\end{figure}

\textbf{Hand-on-Body Prior.} Hands and fingers play an important role as they frequently make contact with the body. However, they have many degrees of freedom, which makes their optimization challenging. Therefore, we learn a hand-on-body prior from 1279 self contact registrations. For this, we use only poses where the minimum point-to-mesh distance between hand and body is $< 1 mm$. These are 718 and 701 poses for the right and left hand, respectively. Since left and right hand are symmetric in SMPL-X, we unite left and right hand poses. Across the 1429 poses, the mean distances per hand vertex to the body surface, $d_m(v_i)$ ranges per vertex from $1.79$ to $5.52$ cm, as visualized in Fig.~\ref{fig:handonbodyprior}. To obtain the weights $h_{v_i}$ in $\mathcal{L}_{H}$, we normalize $d_m(v_i)$ to $[0, 1]$, denoted as $s(d_m(v_i))$, and obtain the vertex weight by $h_{v_i} = -s(d_m(v_i)) + 1$.

\subsection{Mimic-The-Pose (MTP) Data}
\begin{figure}[t]
	\begin{center}
		\includegraphics[width=\linewidth]{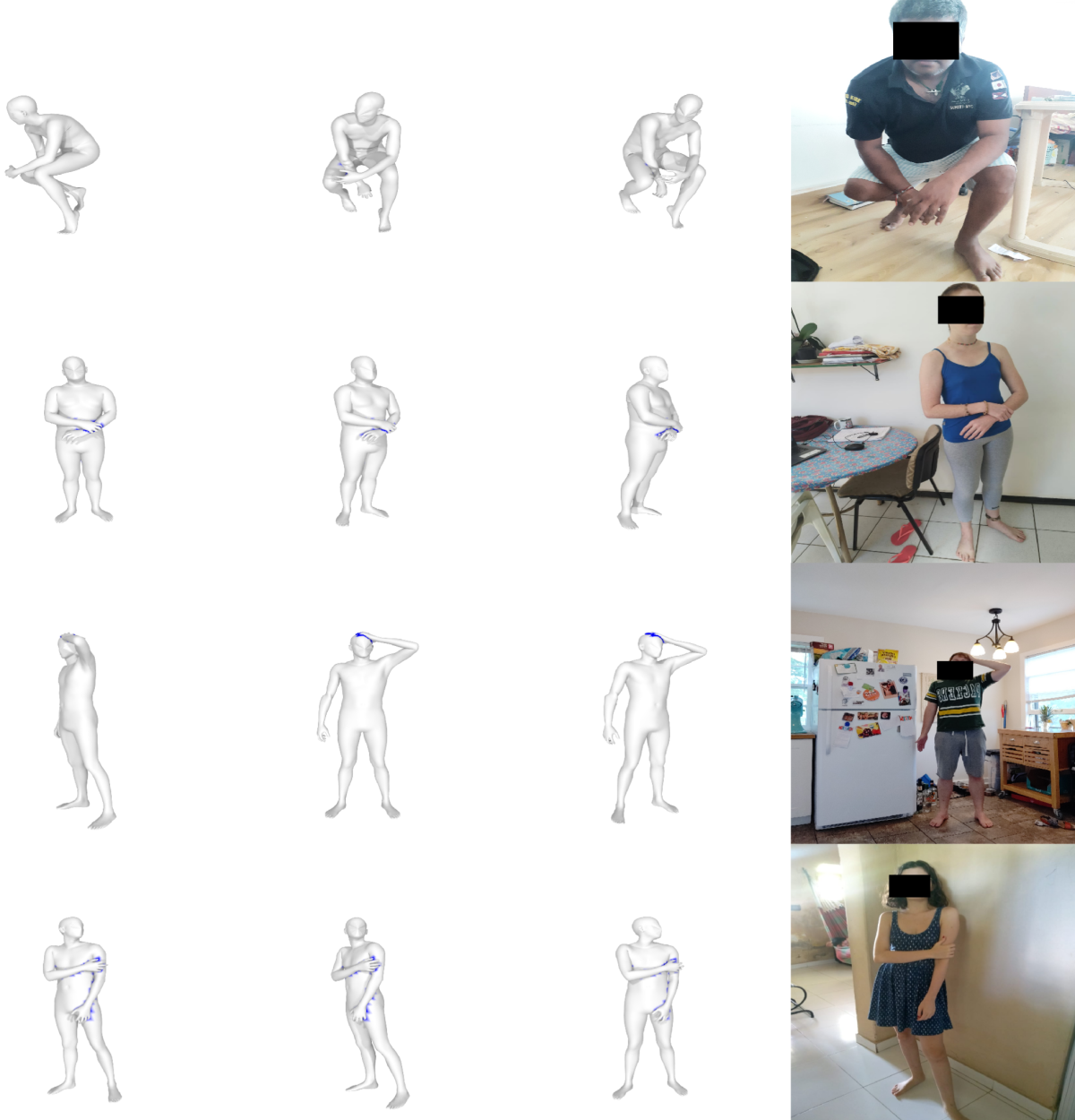}
	\end{center}
	\caption{Presentation format and examples of mimicked poses from the MTP data set. On the left side, the presented pose with contact highlighted in blue. Humans mimicking the poses on the right.}
	\label{fig:canyoumimicthepose}
\end{figure}

\textbf{AMT task details.} It can be challenging to mimic a pose precisely. To simplify the process for workers on AMT, we give detailed instructions, add thumbnails to compare the own image with the presented one and, most importantly, highlight the contact areas; see Fig.~\ref{fig:canyoumimicthepose}. To gain more variety, we also request that participants make small changes in the environment for each image, \eg by  rotating the camera, changing clothes, or turning lights on/off. We also ask participants to mimic the global orientation of the center image. For more variety in global orientation, we vary body roll from $-90^\circ$ to $90^\circ$ in $30^\circ$ steps, resulting in seven different presented global orientations. For example, in the first and third row of Fig.~\ref{fig:canyoumimicthepose}, the center image shows the presented pose from a frontal view. In the second and fourth row, the center body has different orientations. We also ask participants for their height, weight, and gender (M, F, and Non-Binary).

\textbf{\smplifyxmc.} 
In the first stage, we optimize body shape $\beta$ and camera $\Pi$ (focal length, rotation and translation), and body global orientation $\theta_g$, using ground-truth height in meters, $h_{gt}$, and weight in kg, $w_{gt}$. The objective function of the first stage is given as
$$ \mathcal{L}(\beta, \Pi, \theta_g) = \lambda_{\theta_g} \mathcal{L}_{\theta_g} + \lambda_{M} \mathcal{L}_{M} + E_J. $$
$\mathcal{L}_{M} = e^{100|M_h - h_{gt}|} + e^{|M_w - w_{gt}|}$ is the measurements loss, where $M_h$ and $M_w$ are height and weight of mesh $M$. 
We compute height and weight from mesh v-template in a zero pose (T-pose). 
For height, we compute the distance between the top of the head and the mean point between left and right heel. 
For weight, we compute the mesh volume and multiply it by 985 $kg / m^3$, which approximates human body density.
$\mathcal{L}_{\theta_g}$ is a loss that allows rotation around the y-axis, but not around x and z. 

\begin{figure}
	\begin{center}
		\includegraphics[width=\linewidth]{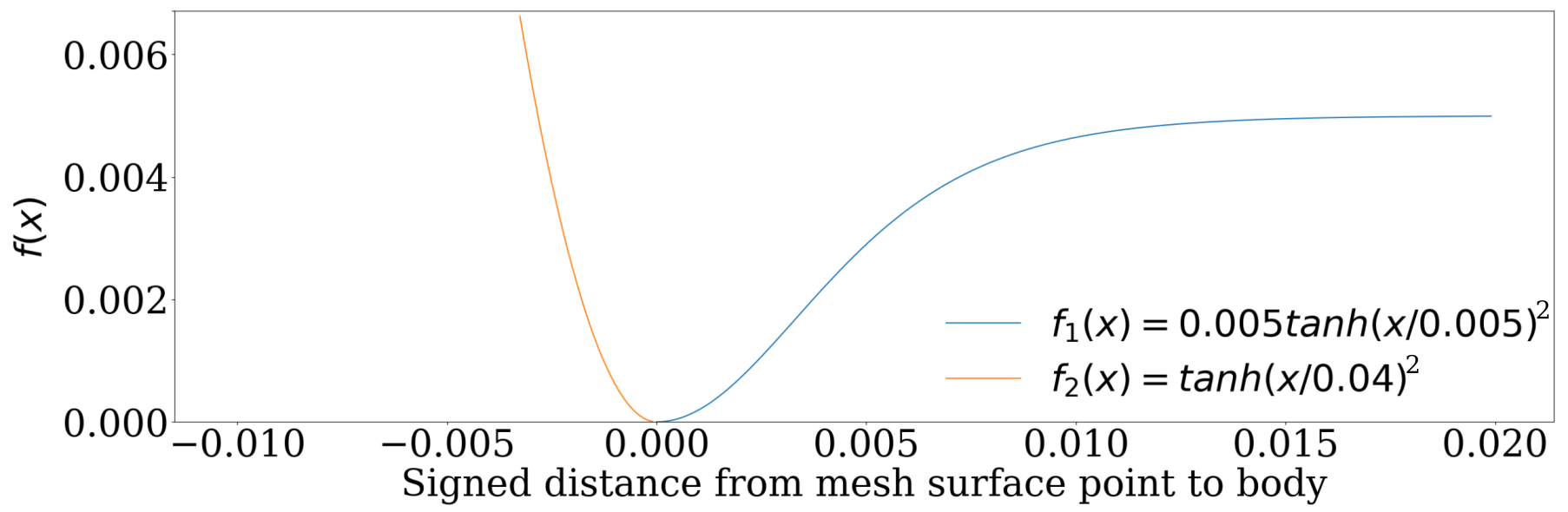}
	\end{center}
	\caption{Functions to regulate the self-contact pushing and pulling term in \smplifyxmc. $f_1$ is used in $\mathcal{L}_{C}$, $f_2$ is used in $\mathcal{L}_{P}$. The parameters ensure that inside vertices are pushed out quickly, while vertices in contact are pulled together as long as they are close enough.}
	\label{fig:tanhfunctions_smplifyxc}
\end{figure}

In Fig.~\ref{fig:tanhfunctions_smplifyxc} we visualize the pushing and pulling terms used in the \smplifyxmc objective. We use 6 PCA components for the hand pose space and initialize the fitting with a mean hand pose. In contrast to SMPLify-X we do not ignore hip joints and double the joint weights for knees and elbows. Before optimization, we resize images and keypoints to a maximum height or width of 500 pixel. Similar to SMPLify-X we use the PyTorch implementation of fast L-BFGS with strong Wolf line search as the optimizer \cite{liu1989limited}. 
We do not use the VPoser pose prior for \smplifyxmc because we have a strong prior from the presented pose.

We notice that the presented global orientation is not always mimicked well. For example, in row 4 of Fig.~\ref{fig:canyoumimicthepose} the presented global orientation has a 60 degree rotation, whereas the mimicked image is taken from a frontal view.
To better initialize the optimization, we select the best body orientation, $\theta_g$, among the seven presented ones based on their re-projection errors; then we compute the camera translation by again minimizing the re-projection error.
We set the initial focal length, $f_x$ and $f_y$, to 2170, which is the average of available EXIF data. 
These values, along with mean shape and presented pose are used to initialize the optimization.

\begin{figure}
\centerline{		\includegraphics[width=\linewidth]{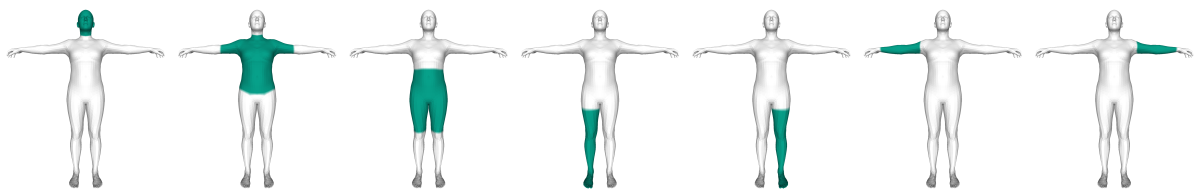}}
\vspace{-0.04in}
	\caption{Body segmented into regions where intersection can happen, since SMPL and SMPL-X are not trained to avoid self intersection. Per segment, we create closed meshes that allow for individual intersection tests. For self-contact, intersections that happen within a segment are not relevant. The hands are not included in any segment, because self intersections within hands or between hands and lower arm are not plausible and need to be resolved.}
	\label{fig:segments}
\end{figure}

In addition, SMPL and SMPL-X have not been trained to avoid self intersection.
Therefore, we identify seven body segments that tend to intersect themselves, e.g.~torso and upper arms (see Fig.~\ref{fig:segments}). We test each segment for self intersection and thereby filter irrelevant intersections from $M_{I}$.

\begin{figure}
\centerline{		\includegraphics[width=\linewidth]{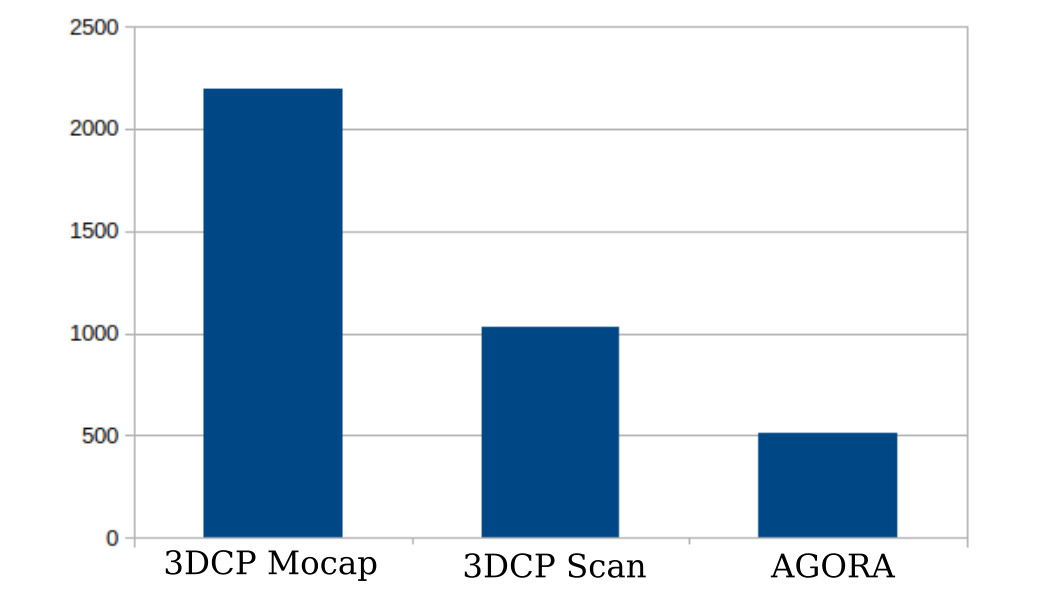}}
\vspace{-0.04in}
	\caption{Image count in MTP Dataset per \threedcp subset.}
	\label{fig:imgsper3dcpsubset}
\end{figure}

\textbf{MTP Dataset Details.} 
We sample meshes from \threedcpscan, \threedcpmocap, and AGORA \cite{patel2021agora} to comprise the presented meshes in \mtp datatset. 
In total, we present 1653 different meshes, from which 1498 (90\%) are contact poses following Definition 3.1 in the main document. 
Of the 1653 meshes, 110 meshes are from \threedcpscan, 1304 meshes are from \threedcpmocap, and 159 are from AGORA. 
We collect at least one image for each mesh.
From the 3731 collected images, 3421 (92\%) images show a person mimicking a contact pose. 
Figure \ref{fig:imgsper3dcpsubset} shows how many image we collected per subset.

\subsection{ Discrete Self-Contact (DSC) Data.} 
\label{subsection:DSC}
\begin{figure}
	\begin{center}
		\includegraphics[width=0.95\linewidth]{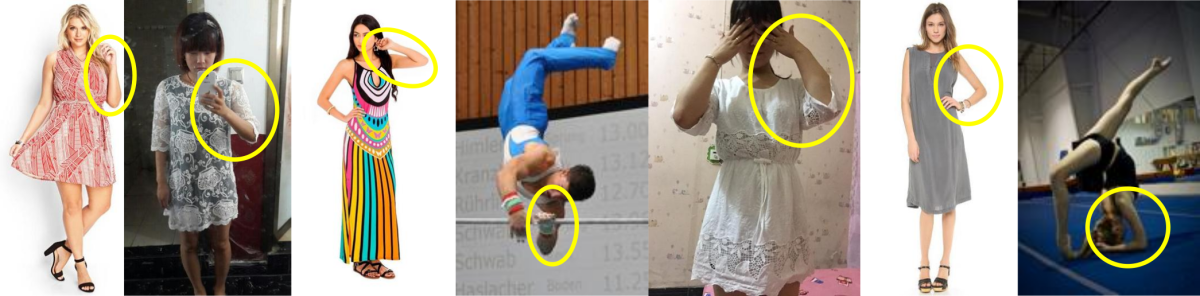}
	\end{center}
	\caption{Discrete self-contact can be challenging to annotate. Here we show a few example images that are annotated as having discrete self-contact between the left upper and lower arm (yellow circle). In the last two images, however, the upper and lower arm are barely touching. We do not consider these to be in self-contact. Another ambiguous case, this time due to occlustion, are the two legs in the first image. An annotator can only assume that the shin and calf are touching, based on semantic knowledge about human pose.}
	\label{fig:DFExamples}
\end{figure}

\textbf{Image selection.} Discrete self-contact annotation may be ambiguous and we find some annotations that we do not consider to be functional self-contact. For example, in Fig.~\ref{fig:DFExamples}, some annotators label the left lower arm and left upper arm to be in contact, because of the slight skin touching at the elbow; we do not treat these as in self-contact. Therefore, we leverage the kinematic tree structure provided by SMPL-X and, in order to train TUCH, ignore the following annotations:
left hand - left lower arm,
left lower arm - left elbow,
left lower arm - left upper arm,
left elbow - left upper arm,
left upper arm - torso,
left foot - left lower leg,
left lower leg - left knee,
left lower leg - left upper leg,
left knee - left upper leg,
right hand - right lower arm,
right lower arm - right elbow,
right lower arm - right upper arm,
right elbow - right upper arm,
right upper arm - torso,
right foot - right lower leg,
right lower leg - right knee,
right lower leg- right upper leg,
right knee - right upper leg.

%% file: supp_sections/supp_05_TUCH.tex
\section{TUCH}

Here we provide details of the \smplifyxmc and \smplifyxdc methods and how we apply them on MTP and DSC data respectively. 

\smplifyxmc is explained in Sec. 4.2 of the main paper. 
It is applied, before the training, to all \mtp images to obtain gender-specific pseudo ground-truth SMPL-X fits. 
To use these fits for TUCH training, two pre-processing steps are necessary. 
First, they are converted to neutral SMPL fits. 
Second, we transform the converted SMPL fits to the camera coordinate frame estimated during \smplifyxmc. 
This is necessary since SPIN assumes an identity camera rotation matrix. 
After that, the data is treated as ground truth during training, which means we apply the regressor loss directly on the converted SMPL pose and shape parameters without in-the-loop fitting.

On the contrary, \smplifyxdc is applied during TUCH training to images with discrete self-contact annotations. 
We run 10 iterations of \smplifyxdc for each image in a mini batch. 

MTP and the DeepFashion subset of DSC do not have ground-truth 2D keypoints but we find OpenPose detections good enough in both cases. 
For the 2D re-projection loss, we use ground-truth keypoints (if available) and OpenPose detections weighted by the detection confidence. Each mini batch consists of 50\% DSC and 50\% MTP data. 

\textbf{Implementation details:} We initialize our regression
network with SPIN weights \cite{kolotouros2019learning}. We use the Adam optimizer \cite{kingma2014adam} and a learning rate of $1e-5$. 

\section{$\text{TUCH}_{\mathit{EX}}$}

One disadvantage of training with fitting in the loop is that it is relatively slow. 
As an alternative, we also explore Exemplar Fine-Tuning (EFT) \cite{joo2020eft}, which is a regression based method for fitting 3D meshes to a single image. 
The fitted SMPL meshes may then be used as pseudo annotations to train a regressor without in-the-loop optimization. 
With this approach, the authors train HMR-EFT, with which they achieve good results on 3DPW and MPI-INF-3DHP.

The idea of using discrete contact annotations is not limited to optimization based approaches. We show that they can also be applied in combination with EFT. Specifically, we extend the regressor loss of EFT with the contact terms from \smplifyxdc. We denote such an ``EFT + contact loss'' approach as EFT-C. Note that the original EFT loss uses a 2D orientation term to match the lower legs orientation, which we do not use here.

Each image in DSC is then paired with a pre-computed pseudo ground truth from EFT-C, and we denote the dataset as [DSC]\textsubscript{EFT-C}. 
Then, we finetune the HMR-EFT network on MTP, [DSC]\textsubscript{EFT-C}, as well as other training data from \cite{joo2020eft}. This new model is called TUCH with EXemplar Finetuning, $\text{TUCH}_{\mathit{EX}}$. 
Unlike TUCH that still performs \smplifyxdc in the training loop, $\text{TUCH}_{\mathit{EX}}$ is supervised only by pre-computed fits so it can be trained faster.

\textbf{Implementation details.} We initialize our network with state-of-the-art HMR-EFT weights. We train $\text{TUCH}_{\mathit{EX}}$ on [COCO-All]\textsubscript{EFT} (CAE), H36M, MPI-INF-3DHP (MI), [DSC]\textsubscript{EFT-C}, and MTP. [COCO-All]\textsubscript{EFT} denotes the COCO dataset after EFT processing, as described in \cite{joo2020eft}. In each batch we use a 10\% CAE, 20\% H36M, 10\% MI, 20\% 3DPW, 20\% [DSC]\textsubscript{EFT-C}, and 20\% MTP. The remaining details are the same as in the TUCH implementation. For the DSC dataset, we only consider images where the full body is visible. To identify these images, we test whether the OpenPose detection confidence of ankles, hips, shoulders, and knees is $\geq$ 0.2. We also ignore discrete contact annotations for connected body parts, as defined in \ref{subsection:DSC}.

%% file: supp_sections/supp_06_Evaluation.tex
\begin{table*}
	\begin{center}
		\setlength\tabcolsep{1pt}
		\setlength{\extrarowheight}{2pt}
		\begin{tabular}{lc|cccc|cccc}
			\toprule[1pt]
			& \multirow{2}{*}{Finetuning Data}	& \multicolumn{4}{c}{MPJPE $\downarrow$}                     & \multicolumn{4}{c}{PA-MPJPE $\downarrow$}                  \\ \cline{3-10} 
			& & contact          & no contact       & unclear     & \underline{total}  & contact          & no contact       & unclear    & \underline{total}   \\ 
			\hline
			{\small HMR-EFT \cite{joo2020eft}}      & -     &     88.3     & 84.6          & 83.6      & 85.3    & 52.1          & \textbf{53.3}          & \textbf{48.5}    &   \textbf{51.7}   \\ 
			\hline
			{\small $\text{TUCH}_{\mathit{EX}}$ }    & \quad {\small CAE + H36M + MI + 3DPW + [DSC]\textsubscript{EFT-C} + MTP}  \quad    & \textbf{82.8} & \textbf{83.2} & \textbf{80.3} & \textbf{82.3} & \textbf{50.4} & 54.1 & 48.7& \textbf{51.7} \\ 
			\bottomrule[1pt]
		\end{tabular}
	\end{center}
	\caption{Evaluation of $\text{TUCH}_{\mathit{EX}}$ for contact classes. CAE = [COCO-All]\textsubscript{EFT} as denoted in \cite{joo2020eft}. Bold numbers indicate the better a result.}
	\label{supptab:results_3DPW_split}
\end{table*}

\begin{figure}
	\begin{center}
		\includegraphics[width=0.95\linewidth]{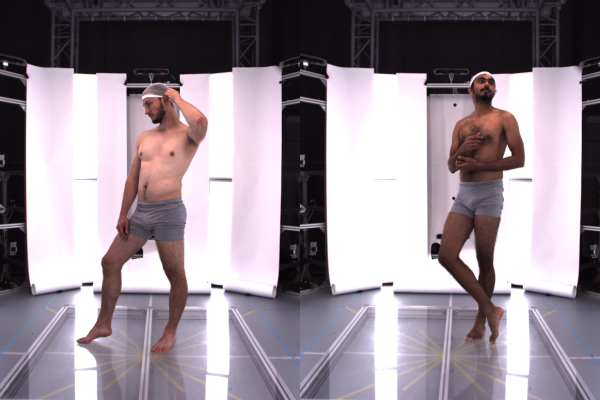}
	\end{center}
	\caption{RGB images from \threedcpscan Scan test set. A subject performing a pose with self-contact in a 3D body scanner.}
	\label{fig:threedcpscanImgs}
\end{figure}
\section{Evaluation}
\textbf{\threedcpscan test images.}
During the scanning process when creating \threedcpscan, we also take RGB photos of subjects being scanned, as shown in Figure \ref{fig:threedcpscanImgs}.
These images have high-fidelity ground-truth poses and shapes from the registration process described in Sec.~\ref{subsec-3dcpscan}, making them a good test set for evaluation purposes.
It is worth noting again that TUCH has never seen these images or subjects, but the contact poses were mimicked in creation of MTP, which is used in training TUCH.

\textbf{TUCH.} 
In Fig.~\ref{fig:qual_result_tuch} we visualize the improvement of TUCH over SPIN qualitatively. One can see that TUCH reconstructs bodies with better self-contact and less interpenetration (row 1 and row 2).
Fig.~\ref{fig:qual_result_tuch2}, on the other hand, shows examples where SPIN is better than TUCH. Four of the images in Fig.~\ref{fig:qual_result_tuch2} do not show the full body (rows 3, 4, 5, and 8). A possible reason why SPIN is better than TUCH in these cases is that MTP images always show the full body of a person, thus TUCH could be more sensitive to occlusion than SPIN. 

We also evaluate the contribution of MTP data by finetuning SPIN only
with it. 
The results are reported in Table \ref{supptab:mtp_only}, where TUCH
(MTP+DSC) is the same as reported in Table 3 of the main paper. This experiment shows
that MTP data alone is already sufficient to significantly improve
state-of-the-art (SOTA) methods on 3DPW benchmarks.
This suggests that the MTP approach is a useful new tool for gathering
data to train neural networks.

\begin{table}
	\begin{center}
		\setlength{\extrarowheight}{2pt}
		\begin{tabular}{lccc}
			\toprule[1pt]
			&   MPJPE $\downarrow$ & PA-MPJPE $\downarrow$ \\ 
			\hline
			SPIN & 96.9 & 59.2 \\
			TUCH (MTP)  &  88.7         & 57.4         \\
			TUCH (MTP+DSC)  &  \textbf{84.9}         & \textbf{55.5}         \\
			\bottomrule[1pt]
		\end{tabular}
	\end{center}
	\caption{Ablation of MTP data and DSC data.}
	\label{supptab:mtp_only}
\end{table}

\textbf{$\text{TUCH}_{\mathit{EX}}$.} 
\begin{table}
	\begin{center}
		\setlength\tabcolsep{2pt}
		\setlength{\extrarowheight}{2pt}
		\begin{tabular}{lcccc}
			\toprule[1pt]
			& \multicolumn{2}{c}{MPJPE $\downarrow$}     & \multicolumn{2}{c}{PA-MPJPE $\downarrow$}  \\  \cline{2-5} 
			& 3DPW & MI        & 3DPW & MI      \\   
			\hline
			HMR-EFT \cite{joo2020eft}        & 85.3          & 105.3               & \textbf{51.7}          & 68.4          \\
			$\text{TUCH}_{\mathit{EX}}$  & \textbf{82.3} & \textbf{101.5}          & \textbf{51.7} & \textbf{66.4}         \\
			\bottomrule[1pt]
		\end{tabular}
	\end{center}
	\caption{Training with 3DPW. Evaluation on 3DPW and MPI-INF-3DHP (MI). We report MPJPE and PA-MPJPE for different subsets of our data set.}
	\label{supptab:results_w3dpw}
\end{table}
\begin{table}
	\begin{center}
		\setlength\tabcolsep{2pt}
		\setlength{\extrarowheight}{2pt}
		\begin{tabular}{lccc}
			\toprule[1pt]
		               & MPJPE $\downarrow$ & PA-MPJPE $\downarrow$ & MV2VE $\downarrow$ \\
		    \hline
			      HMR-EFT \cite{joo2020eft}   & \textbf{71.4}           & 48.3              & 83.9  \\
			$\text{TUCH}_{\mathit{EX}}$   & 73.4           & \textbf{43.3}              & \textbf{82.8}       \\
			\bottomrule[1pt]  
		\end{tabular}
	\end{center}
	\caption{Evaluation on \threedcpscan test images. We report MPJPE, PA-MPJPE, and MV2VE. }
	\label{supptab:threedcpscanResult}
\end{table}

For an additional comparison with SOTA EFT
\cite{joo2020eft}, we evaluate our $\text{TUCH}_{\mathit{EX}}$ model
on the same datasets (3DPW, MPI-INF-3DHP (MI), and \threedcpscan) with error
measures (MPJPE, PA-MPJPE, and MV2VE) like TUCH, see Tables
\ref{supptab:results_3DPW_split}, \ref{supptab:results_w3dpw}, and
\ref{supptab:threedcpscanResult}. 

The MPJPE of $\text{TUCH}_{\mathit{EX}}$
improves over HMR-EFT when evaluated on 3DPW. PA-MPJPE improves for
contact poses and is overall on-par. Also the results on MPI-INF-3DHP
improve. For the \threedcpscan test set, PA-MPJPE improves. This shows that our 
data can not only be used with optimization based approaches, but also 
with exemplar fine-tuning, and that it allows us to improve the latest models in terms of estimating poses with contact.

%% file: supp_sections/supp_06_Evaluation_Qualitative.tex
\begin{figure*}[h]
	\begin{center}
		\includegraphics[trim={0 1cm 0 0.1cm},clip,width=\linewidth]{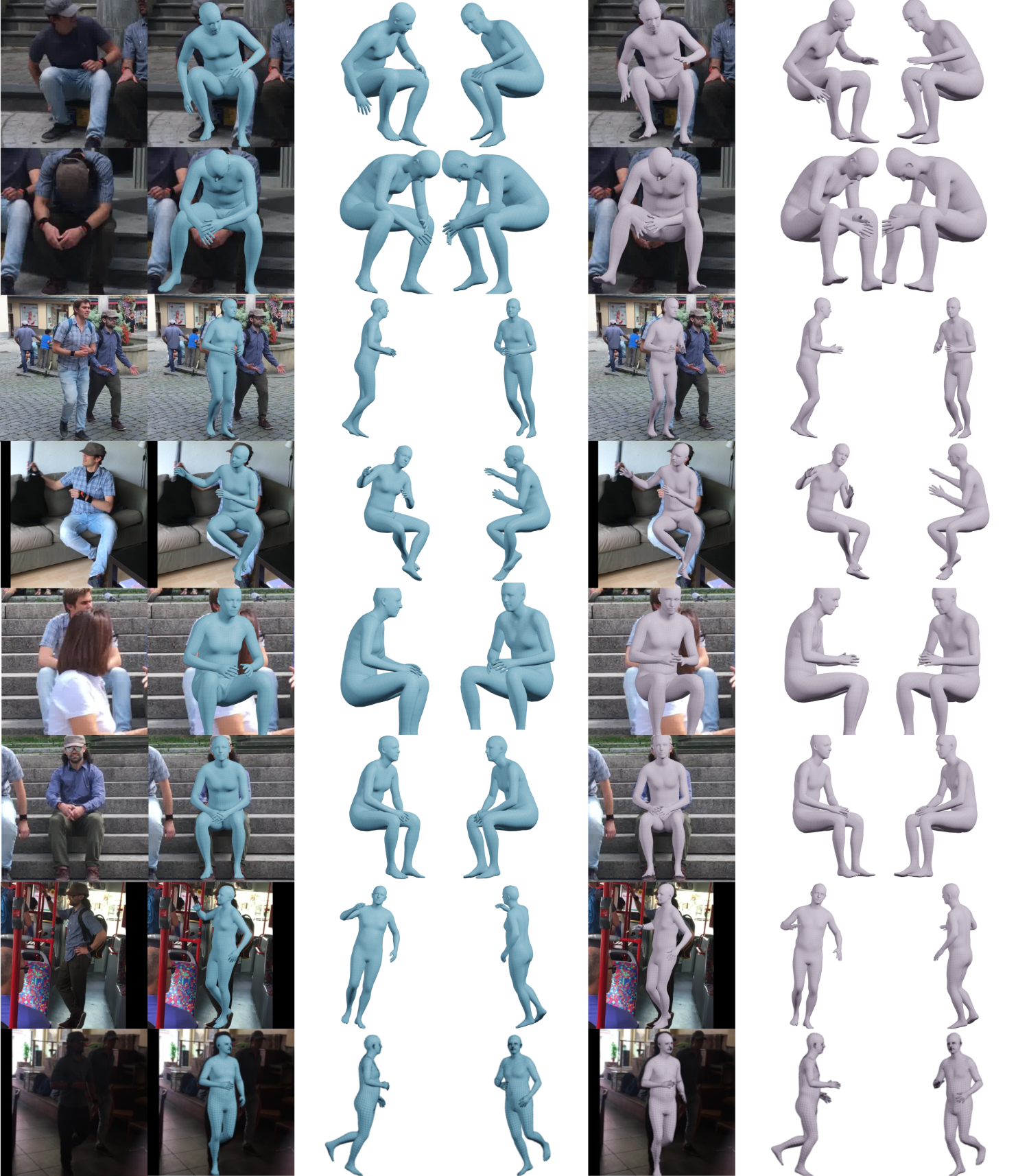}
	\end{center}
	\caption{Qualitative results on the self-contact subset of 3DPW. We find all images with an improvement on MPJPE and PA-MPJPE $\geq$ 10 mm. From this subset, we select interesting poses. Left column, RGB image for reference. In blue, TUCH result and in violet, the SPIN result.}
	\label{fig:qual_result_tuch}
\end{figure*}

\begin{figure*}[h]
	\begin{center}
		\includegraphics[width=0.99\linewidth]{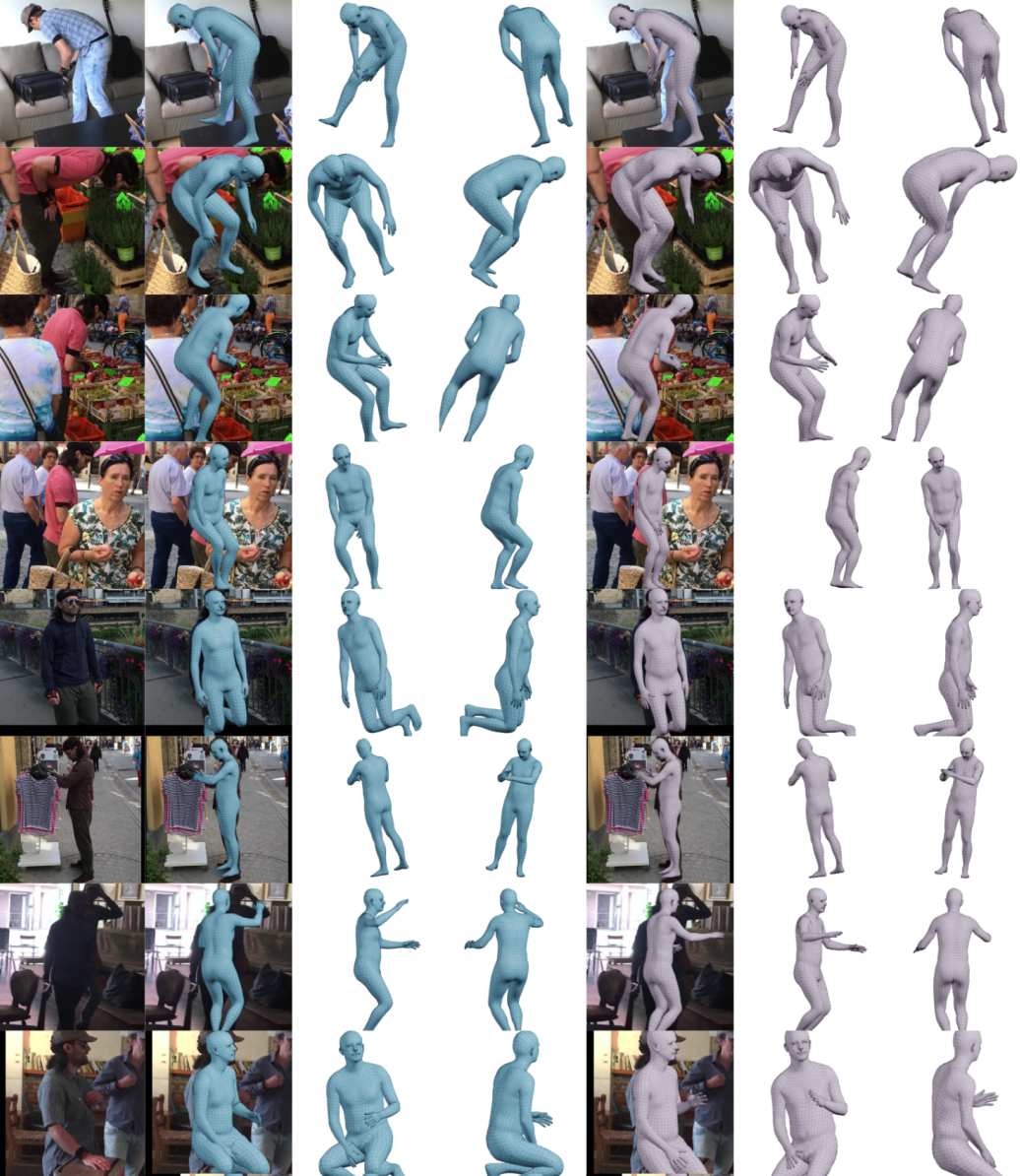}
	\end{center}
	\caption{Qualitative results on the self-contact subset of 3DPW. We find all images where SPIN is better than TUCH by at least 10 mm for MPJPE and PA-MPJPE. From this subset, we select interesting poses. Left column, RGB image for reference. In blue, TUCH result and in violet, the SPIN result.}
	\label{fig:qual_result_tuch2}
\end{figure*}